\documentclass{tut2018}

\input{preamble}

\RemoveLogos

\usepackage[colorlinks=true,breaklinks=true,bookmarks=false,urlcolor=blue,%
  citecolor=blue,linkcolor=blue,bookmarksopen=false,draft=false]{hyperref}

\usepackage[square,numbers]{natbib}

\usepackage{comment,dsfont,mathtools,bm,xspace}
 
\allowdisplaybreaks

\DeclareMathOperator{\ind}{\mathds{1}}  

\newcommand{\mycommentbegin}{}

\newcommand{\exlim}[2]{\mathop\mathbb{E}\limits_{#1}\left[#2\right]}

\newcommand{\prnbig}[1]{\big({#1}\big)} 
\newcommand{\KL}[2]{\mathsf{KL}\prnbig{{#1}\,\|\,{#2}}}

\newcommand{\innprod}[1]{\big\langle{#1} \big\rangle} 
\newcommand{\norm}[1]{\left\|{#1}\right\|} 
\newcommand{\dnorm}{\mathcal{\pi}_{\text{cal}}}

\newcommand{\pihatstar}{\widehat{\pi}^\star}

\newcommand{\ceta}{c_\eta}
\newcommand{\pib}{\pi^{\mathsf{b}}}
\newcommand{\tmix}{t_{\mathsf{mix}}}
\newcommand{\mumin}{\mu_{\mathsf{min}}}

\newcommand{\cA}{{\mathcal{A}}}
\newcommand{\cP}{{\mathcal{P}}}

\newcommand{\cS}{{\mathcal{S}}}
\newcommand{\cM}{{\mathcal{M}}}

\newcommand{\cT}{{\mathcal{T}}}
\newcommand{\cU}{{\mathcal{U}}}

\newcommand{\defn}{\coloneqq}

\newcommand{\taumax}{\tau_{\max}}

\newcommand{\myrho}{d^{\mathsf{b}}}

\newcommand{\mymid}{\,|\,}
\newcommand{\cb}{c_{\mathsf{b}}}

\newcommand{\LCBQ}{{\sf LCB-Q}\xspace}

\newcommand{\Cstar}{C^{\star}}

\newcommand{\rob}{{\mathsf{rob}}}
\newcommand{\TV}{{\mathsf{TV}}}

\newcommand{\no}{0} 
\newcommand{\ror}{\sigma} 
\newcommand{\unb}{\cU} 

\newcommand{\Tpess}{\widehat{\mathcal{T}}_{\mathsf{pe}}}

\usepackage[linesnumbered,ruled,vlined]{algorithm2e}
\usepackage[noend]{algorithmic}

\definecolor{yxc}{RGB}{255,0,0}
\definecolor{yjc}{RGB}{25,25,255}
\definecolor{ytw}{RGB}{255,105,180}

\begin{document}

\CHAPTERNO{\phantom{Chapter 1}}
\DOI{}

\TITLE{Statistical and Algorithmic Foundations of Reinforcement Learning}


\AUBLOCK{

\AUTHOR{Yuejie Chi}
\AFF{Department of Statistics and Data Science, Yale University, New Haven, CT 06511, USA,
\EMAIL{yuejie.chi@yale.edu}}
\AUTHOR{Yuxin Chen, Yuting Wei}
\AFF{Department of Statistics and Data Science, University of Pennsylvania, Philadelphia, PA 19104, USA,
\textbraceleft\EMAIL{yuxinc@wharton.upenn.edu}, \EMAIL{ytwei@wharton.upenn.edu}\textbraceright}

}

\CHAPTERHEAD{}

\ABSTRACT{As a paradigm for sequential decision making in unknown environments, reinforcement learning (RL) has received a flurry of attention in recent years. However, the explosion of model complexity in emerging applications and the presence of nonconvexity exacerbate the challenge of achieving efficient RL in sample-starved situations, where data collection is expensive, time-consuming, or even high-stakes (e.g., in clinical trials, autonomous systems, and online advertising). How to understand and enhance the sample and computational efficacies of RL algorithms is thus of great interest. In this tutorial, we aim to introduce several important algorithmic and theoretical developments in RL, highlighting the connections between new ideas and classical topics. Employing Markov Decision Processes as the central mathematical model, we cover several distinctive RL scenarios (i.e., RL with a simulator, online RL, offline RL, robust RL, and RL with human feedback), and present several mainstream RL approaches (i.e., model-based approach, value-based approach, and policy optimization). Our discussions gravitate around the issues of sample complexity, computational efficiency, as well as algorithm-dependent and information-theoretic lower bounds from a non-asymptotic viewpoint.  
}


\maketitle
 
\tableofcontents

\section{Introduction}

Reinforcement learning (RL) \citep{sutton2018reinforcement,szepesvari2010algorithms}, which is frequently modeled as learning and decision making in a Markov decision process (MDP), is garnering growing interest in recent years due to its remarkable success in practice. A core objective of RL is to search for a policy---based on a collection of noisy data samples---that approximately maximizes the expected cumulative rewards in an MDP, without direct access to a precise description of the underlying model.\footnote{Here and throughout, the ``model'' refers to the transition kernel and the rewards of the MDP taken collectively.}    
In contemporary applications, it is increasingly more common to encounter environments with prohibitively large state and action space, thus exacerbating the challenge of collecting abundant samples, when data collection is expensive, time-consuming, or even high-stake (such as clinical trials, online advertisements, autonomous systems, to name just a few). To enable faithful policy learning in the sample-starved regime (i.e., the regime where the model complexity overwhelms the sample size), it is crucial to obtain a quantitative picture of the fundamental trade-off between sample complexity and statistical accuracy, and to design efficient algorithms that provably achieve the optimal trade-off.

Broadly speaking, there are at least two common algorithmic approaches: a model-based approach and a model-free one. The model-based approach decouples model estimation and policy learning tasks; more specifically, one first estimates the unknown model using the data samples in hand, and then leverages the fitted model to 
perform planning---a task that aims to find the optimal policy for the prescribed model and can be accomplished by resorting to Bellman's principle of optimality \citep{bellman1952theory}. A notable advantage of model-based algorithms is their flexibility: the learned model can be adapted to perform new ad-hoc tasks without revisiting the data samples.  In comparison, the model-free approach attempts to directly interact with the environment---in the form of a policy that selects actions based on perceived states of the environment---from the collected data samples without learning the model explicitly. 
Therefore, model-free algorithms are able to adapt to the model changes on the fly and are often memory-efficient. For example, Q-learning \citep{watkins1992q}, which seeks to learn the optimal action-value function by observing what happens under a behavior policy, is a widely used model-free approach. 
Characterizing the sample efficiency of both approaches has been the focal point of a large body of recent works.

%
%

Another prevalent class of model-free algorithms is the policy-based approach (or policy optimization), prominent examples including the policy gradient (PG) method \citep{williams1992simple,sutton1999policy}. On a high level, 
PG methods follow an optimizer's perspective by formulating a value maximization problem with regard to parameterized policies, and and perform gradient updates to improve the policy estimates iteratively---often based on noisy feedback received from the environments. PG methods and their variants have become the de facto standard practices in an increasing number of domains, due to their seamless integration with neural network parameterization and adaptivity to various problem setups involving discrete, continuous or mixed action and state spaces.
Despite the enormous empirical success of PG methods, little was known about their theoretical convergence properties---especially when it comes to finite-time global convergence---due to the notorious nonconcavity of value functions. Until very recently, such understandings began to emerge, leveraging the fact that PG methods typically operate on highly structured model classes, whose induced optimization landscapes turn out to be more benign and tractable than previously believed. The execution of problem-dependent tailored analyses, rather than relying on black-box optimization theory, has fueled recent breakthroughs pioneered by \citet{fazel2018global,agarwal2019optimality,bhandari2024global}, to name just a few.

An explosion of research has been conducted over the past few years towards advancing the frontiers of the above-mentioned topics, leveraging toolkits that lie at the heart of information and data science, such as nonconvex optimization, high-dimensional statistics, information theory,  and machine learning theory. In this tutorial, our goal is not merely to cover the fundamentals of RL, but also highlight cutting-edge algorithmic ideas and distinctive paradigms.

\paragraph{Notation.} Before proceeding, let us introduce a couple of notation that shall be used throughout this tutorial. 
For any set $\cS=\{1,2,\dots,S\}$, we denote by $\Delta(\cS)$ the set of all distributions over $\cS$.  
%
Let $\mathcal{X}\coloneqq \big( |\mathcal{S}|,|\mathcal{A}|, \frac{1}{1-\gamma}, H, \frac{1}{\varepsilon} \big)$.  The notation $f(\mathcal{X}) = O(g(\mathcal{X}))$   means there exists a universal constant $C_1>0$ such that $f\leq C_1 g$, whereas the notation $f(\mathcal{X}) = \Omega(g(\mathcal{X}))$ means $g(\mathcal{X}) = O(f(\mathcal{X}))$. In addition, the notation $\widetilde{O}(\cdot)$ (resp.~$\widetilde{\Omega}(\cdot)$) is defined in the same way as ${O}(\cdot)$ (resp.~$\Omega(\cdot)$) except that it ignores logarithmic factors. Let $\ind\{\cdot\}$ denote the indicator function. For any probability vector $q\in \mathbb{R}^{S}$  and any vector $V \in \mathbb{R}^S$,   define the variance of $V$ w.r.t.~the distribution $q$ as
\begin{align}
	\mathsf{Var}_{q} (V)\coloneqq \sum_i q_iV_i^2 - \Big(\sum_i q_i V_i \Big)^2. 
	\label{eq:defn-Var-P-V}
\end{align}

\section{Preliminaries}
\label{chap:mdp}

Before proceeding to concrete RL results, we begin by introducing some preliminaries of MDPs and dynamic programming. 

\subsection{Markov decision processes}

Markov decision processes (MDPs) form the prototypical models for the analysis of RL algorithms. 
We shall only present some preliminary concepts and basic notation, and more in-depth discussions can be found in classical textbooks \citep{puterman2014markov,bertsekas2017dynamic}.

\paragraph{Discounted infinite-horizon MDPs.}  
For ease of presentation, a large fraction of the results presented in this tutorial (except for online RL) are concerned with (staionary) discounted infinite-horizon MDPs, which we introduce here. 
Denote by $\mathcal{M} = \{\cS, \cA, P, \gamma, r\}$ a discounted infinite-horizon MDP, which comprises the following key components: 
(i) $\cS = \{1,2,\cdots, S\}$: a finite state space containing $S$ different states; (ii) $\cA = \{1,2,\cdots, A\}$:  an action space containing $A$ different actions; (iii) $P: \cS\times \cA \rightarrow \Delta(\cS)$: the transition probability kernel of the MDP, so that $P (\cdot \mymid s,a)$ represents the transition probability from state $s$ when action $a$ is executed; (iv) $\gamma \in [0,1)$: the discount factor, with $\frac{1}{1-\gamma}$ representing the effective horizon; (v) $r: \cS\times \cA \rightarrow [0,1]$: the deterministic reward function (namely, $r(s,a)$ indicates the immediate reward received when action $a$ is taken in state $s$). Without loss of generality, we assume throughout that the immediate rewards are normalized, so that they all lie within the interval $[0,1]$. 
Additionally, we find it helpful to introduce the convenient notation below: 
\begin{align}
	P_{s,a} \coloneqq P(\cdot \mymid s,a) \in \mathbb{R}^{1\times S}.
\end{align}

\begin{itemize}

\item {\em Policy.}
A stationary policy $\pi: \cS \rightarrow \Delta(\cA)$ is a possibly randomized action selection rule executed by the agent; that is, $\pi(a \mymid s)$ represents the probability of choosing $a$ in state $s$. When $\pi$ is a deterministic policy, we abuse the notation by letting $\pi(s)$ represent the action chosen by the policy $\pi$ in state $s$. 

	\item {\em Value function and Q-function.}
		To evaluate the goodness of a policy, we introduce the value function $V^\pi$ and Q-function $Q^\pi$ associated with policy $\pi$. Specifically, the value function $V^\pi: \cS \rightarrow \mathbb{R}$ of policy $\pi$ is defined as the expected discounted cumulative reward as follows:
\begin{align}
	\forall s\in\cS: \quad 
	V^\pi(s) \coloneqq \mathbb{E} \left[\sum_{t=0}^\infty \gamma^t r(s_t, a_t) \mid s_0 = s \right],
	\label{eq:defn-Vpi-inf}
\end{align}
where the expectation is taken over the sample trajectory $\{(s_t, a_t)\}_{t\geq 0}$ generated in a way that $a_t \sim \pi(\cdot \mymid s_t)$ and $s_{t+1} \sim P(\cdot \mymid s_t, a_t)$ for all $t\geq 0$ from an initial action $s_0 = s$. Given that all immediate rewards lie within $[0,1]$, it is easily verified that $0\leq V^\pi(s) \leq \frac{1}{1-\gamma}$ for any policy $\pi$. The Q-function (or action-state function) of policy $\pi$ can be defined analogously as follows:
\begin{align}
	\forall (s,a)\in\cS \times \cA: \quad 
	Q^\pi(s,a) \coloneqq \mathbb{E} \left[\sum_{t=0}^\infty \gamma^t r(s_t, a_t) \mid s_0 = s, a_0 = a \right],
\end{align}
which differs from \eqref{eq:defn-Vpi-inf} in that it is also conditioned on the initial action $a_0=a$. 
Additionally, 
if the initial state is randomly drawn from a given state distribution $\rho \in \Delta(\cS)$, then we can define the weighted value function of policy $\pi$ as follows:
\begin{align}
	V^\pi(\rho) \coloneqq \mathop{\mathbb{E}}\limits_{s\sim \rho}\big[V^\pi(s)\big].
\end{align}

	\item {\em Optimal policy, value and Q-functions.} It turns out that there exists at least one deterministic policy---denoted by $\pi^{\star}$---that can simultaneously maximize the value function $V^{\pi}(s)$ and Q-function $Q^{\pi}(s,a)$ for simultaneously for all $(s,a)\in \cS\times \cA$ (see, e.g., \cite{puterman2014markov}). We shall refer to this policy as the optimal policy. The resulting optimal value function and Q-function are denoted by
\begin{align}
	V^{\star}(s) =V^{\pi^{\star}}(s) = \max_{\pi} V^{\pi}(s) \qquad \mbox{and}\qquad   Q^{\star}(s,a)  =Q^{\pi^{\star}}(s,a)  = \max_{\pi} Q^{\pi}(s,a) .
\end{align}

	\item {\em Discounted occupancy distributions.}
We also find it convenient to introduce the {\em discounted occupancy distributions} associated with policy $\pi$ as follows:  
%
\begin{align}
	\forall (s,a)\in \cS \times \cA: \quad d_\rho^\pi(s,a) &\coloneqq (1-\gamma) \sum_{t=0}^\infty \gamma^t \mathbb{P}(s_t = s, a_t =a \mid s_0 \sim \rho), \label{eq:visitation_dis_sa}
\end{align}
where the randomness is over a sample trajectory that starts from an initial state $s_0 \sim \rho$ and that follows policy $\pi$ (i.e., $a_t \sim \pi(\cdot \mymid s_t)$ and $s_{t+1} \sim P(\cdot \mymid s_t, a_t)$ for all $t\geq 0$).

%

\end{itemize}

\paragraph{Finite-horizon MDPs.}  
Let us also introduce the basics of time-inhomogeneous or non-stationary finite-horizon MDPs, which will be considered in Section~\ref{chapter:online-RL} concerning online RL.  
Denote by $\mathcal{M}=(\mathcal{S},\mathcal{A},H, \{P_h\}_{h=1}^H, \{r_h\}_{h=1}^H)$ a finite-horizon MDP, where $\mathcal{S}\coloneqq \{1,\cdots S\}$ (resp.~$\mathcal{A}\coloneqq \{1,\cdots, A\}$) is the state space (resp.~action space) as before.  
We let $H$ denote the horizon length, and $P_h : \cS \times \cA \rightarrow \Delta (\cS) $ (resp.~$r_h: \cS \times \cA \rightarrow [0,1]$)
represents the probability transition kernel (resp.~reward function) at the $h$-th time step, $1\leq h\leq H$, respectively. 
More specifically, $P_h(\cdot\mymid s,a)\in \Delta(\cS)$ stands for the transition probability vector from state $s$ at time step $h$ when action $a$ is taken, 
while $r_h(s,a)$ indicates the immediate reward received at time step $h$ for a state-action pair $(s,a)$ (which is assumed to be deterministic and falls within the range $[0,1]$). 

A deterministic policy is represented by $\pi =\{\pi_h\}_{h=1}^H$ with $\pi_h: \mathcal{S} \rightarrow \mathcal{A}$ the action selection rule at time step $h$, so that 
$\pi_h(s)$ specifies which action to execute in state $s$ at time step $h$.  
The value function $V^{\pi}_{h}(s)$ of a (deterministic) policy $\pi$ at step $h$ is defined as the expected cumulative rewards received between time steps $h$ and $H$ when executing this policy from an initial state $s$ at time step $h$, namely, 
\begin{align}
	\label{eq:def_Vh}
	V^{\pi}_{h}(s) & \coloneqq  \mathop{\mathbb{E}}
	\left[  \sum_{t=h}^{H} r_{t}\big(s_{t},\pi_{t}(s_{t}) \big) \,\Big|\, s_{h}=s \right], 
\end{align}
where the expectation is taken over the randomness of the MDP trajectory $\{s_t \mid h\leq t\leq H\}$. 
The  Q-function $Q^{\pi}_h(s,a)$ of a policy $\pi$ at step $h$ can be defined analogously 
except that the action at step $h$ is fixed to be $a$, that is, 
\begin{align} 
	\label{eq:def_Qh}
	Q^{\pi}_{h}(s,a) & \coloneqq r_{h}(s,a)+ \mathop{\mathbb{E}}
	\left[  \sum_{t=h +1}^{H} r_t \big(s_t,\pi_t(s_t) \big) \,\Big|\, s_{h}=s, a_h  = a\right].
\end{align}
In addition, we define $V^{\pi}_{H+1}(s)= Q^{\pi}_{H+1}(s,a)=0$ for any policy $\pi$ and any state-action pair $(s,a)\in \cS \times \cA$. 
%
A policy $\pi^{\star} =\{\pi_h^{\star}\}_{h=1}^H$ is said to be an optimal policy if it maximizes the value function simultaneously for all states among all policies. 
The resulting optimal value function $V^{\star} =\{ V_h^{\star} \}_{h=1}^H $ and optimal Q-functions $Q^{\star} =\{ Q_h^{\star} \}_{h=1}^H $ satisfy  
\begin{align}
	V_h^{\star}(s) =V_h^{\pi^{\star}}(s) = \max_{\pi} V_h^{\pi}(s) \qquad \mbox{and}\qquad   Q_h^{\star}(s,a)  =Q_h^{\pi^{\star}}(s,a)  = \max_{\pi} Q_h^{\pi}(s,a) 
\end{align}
for any $(s,a,h)\in \cS\times \cA \times [H]$. 

\begin{remark}
There are a few other popular MDP models widely considered in the literature, such as the average-reward infinite-horizon model \cite{wan2021learning,zurek2024spanbased}, and the time-homogeneous finite-horizon model \cite{ren2021nearly,zhang2022horizon}, each of which shares some similarities with the ones under exposition here, but also presents unique challenges. We refer the interested reader to these papers and the references cited therein for further readings.
 
\end{remark}
%

\subsection{Dynamic programming}
\label{sec:dp}

Next, we present some basics about dynamic programming \citep{bertsekas2017dynamic}. 
Of crucial importance in the development are the Bellman operators \citep{bellman1952theory}. In what follows, we introduce the Bellman optimality operator for discounted infinite-horizon MDPs; the finite-horizon counterpart can be found in classical dynamic programming books \citep{puterman2014markov,bertsekas2017dynamic}. 

%

\paragraph{Bellman optimality operator.} The Bellman optimality operator $\mathcal{T}$, which is a mapping from  $\mathbb{R}^{S \times A}$ to itself, is defined such that the $(s,a)$-th entry of $\mathcal{T}(Q)$ is given by
\begin{equation}\label{eq:bellman_optimality_operator}
	\mathcal{T}(Q)(s,a) := r(s,a) + \gamma \mathop{\mathbb{E}}\limits_{s^{\prime} \sim P(\cdot| s,a)}  \Big[ \max_{a^{\prime}\in \cA} Q(s^{\prime}, a^{\prime}) \Big].
\end{equation}
The optimal Q-function $Q^{\star}$ is the unique fixed point of the Bellman operator, namely, 
\begin{equation}\label{eq:bellman_fix_point}
  Q^{\star} = \mathcal{T}(Q^{\star}).
\end{equation}
An important property of the Bellman optimality operator is $\gamma$-contraction, that is, for any $Q_1, Q_2\in\mathbb{R}^{S\times A}$, it holds that
\begin{align}
\| \mathcal{T}(Q_1) -\mathcal{T}(Q_2) \|_{\infty} & \leq \gamma  \|  Q_1 - Q_2 \|_{\infty} .
\end{align}



\paragraph{Policy iteration and value iteration.} We now describe two efficient algorithms to compute the optimal policy $\pi^{\star}$ and the optimal Q-function $Q^{\star}$.
\begin{itemize}

	\item {\em Policy iteration.} Initialize the policy iterate to $\pi_0$. In each iteration $k=1,\ldots$, evaluate $Q^{k} = Q^{\pi_{k-1}}$ and update the policy as
\begin{equation} \label{eq:policy_iteration}
\pi_k (a|s) = \argmax_{a \in \cA}\; Q^{k} (s,a).
\end{equation}
In other words, policy iteration updates a policy by selecting the action ``greedily'' w.r.t.~the Q-function of the previous policy iterate.
 
\item {\em (Q)-value iteration.} Initialize the Q-estimate to $Q_0$. For $k=1,\ldots$, update the Q-function as
\begin{equation} \label{eq:value_iteration}
Q_k = \mathcal{T}(Q_{k-1}).
\end{equation}
In other words, value iteration updates the Q-function by iteratively applying the Bellman optimality operator to the previous Q-function iterate.
\end{itemize}
Both approaches provably achieve fast linear convergence to the optimal Q-function, as stated below.
\begin{lemma}[Linear convergence of policy/value iteration]
For both policy iteration and value iteration, one has
$ \| Q_k - Q^{\star}\|_{\infty} \leq \gamma^k  \| Q_o - Q^{\star}\|_{\infty} $
for all $k \geq 1$.
\end{lemma} 

Hence, to achieve $\| Q_k - Q^{\star}\|_{\infty}\leq \varepsilon $, it takes no more than
$  \frac{1}{1-\gamma}  \log \big( \frac{\| Q_{0} - Q^{\star}\|_{\infty}}{\varepsilon }\big) $
iterations, which depends linearly on the so-called effective horizon $\frac{1}{1-\gamma} $  of the discounted MDP.

\section{Reinforcement learning with a generative model}
\label{chapter:generative-model}

In this section, we discuss the sample complexity of RL when samples are drawn from an idealized sampling mechanism, known as the generative model or the simulator.

\subsection{Sampling from a generative model} 
For concreteness, consider the discounted infinite-horizon MDP throughout this section. 
Initially studied by \cite{kearns1999finite}, 
a generative model---also referred to as a simulator---allows one to query any state-action pair and sample an independent state transition from the ground-truth transition kernel. More precisely, each query to the generative model with a pair $(s,a)\in \mathcal{S}\times \mathcal{A}$ returns an independent sample: 
\begin{align}
    \label{eq:generative-model-homogeneous}
    s'  \overset{\mathrm{ind.}}{\sim} P(\cdot \mymid s,a).  
\end{align}
Studying RL with a generative model offers several key advantages.  To begin with,  simulators have been developed in several applications to simulate complex, real-world environments. Perhaps more importantly, even for scenarios where simulators are unavailable, investigating RL under this idealistic sampling mechanism helps us distill core design principles, establish theoretical benchmarks, and understand the fundamental statistical limits of RL algorithms. These insights might in turn guide us towards more practical and robust solutions.  

In this section, we concentrate on the task of policy learning, 
which seeks to identify a policy that approximately maximizes the value function based on the data samples at hand. 
To be precise, for any target accuracy level $\varepsilon>0$, the aim of policy learning is to compute an $\varepsilon$-accurate policy $\widehat{\pi}$ obeying
\begin{align}
	\forall (s,a)\in\cS\times \cA: \qquad V^{\widehat{\pi}}(s) \geq V^{\star}(s) -\varepsilon.
\end{align}
Recall that for the normalized reward setting with $0\leq r\leq 1$, the value function falls within the range $[0,\frac{1}{1-\gamma}]$; this means that the  range of the target accuracy level $\varepsilon$ should be $\varepsilon \in [0, \frac{1}{1-\gamma}]$. 
Naturally, one would hope to achieve policy learning using as few samples as possible. This task has been extensively studied (e.g., \cite{kearns1999finite, kakade2003sample, azar2013minimax,wang2017randomized,sidford2018near,sidford2018variance,li2024q,wainwright2019stochastic,wainwright2019variance,yang2019sample,pananjady2019value,khamaru2021temporal,yin2021near,jin2021towards,wang2021sample,li2022minimax,jin2024truncated,zurek2024spanbased,zurek2024plug}), with \cite{azar2013minimax,sidford2018near,agarwal2020model,li2024breaking,zurek2024plug} achieving asymptotic optimality in different regimes.  
There are at least two prominent approaches that are widely used in practice: the model-based approach, and the model-free approach, to be detailed next.

\subsection{The model-based approach}
\label{sec:model-based-policy-eval-sim}

The class of model-based algorithms typically starts with model estimation---namely, constructing an empirical MDP $\widehat{\mathcal{M}}$ based on all collected data samples---and then ``plug in'' the estimated model directly into the Bellman recursion to derive a policy estimate (using classical dynamic programming algorithms like  Q-value iteration and policy iteration \citep{bertsekas2017dynamic} discussed in Section~\ref{sec:dp}).

\subsubsection{Algorithms}
To be precise, let us describe below the most natural version of the model-based algorithms. 
\begin{itemize}
	\item[1)] {\em Model estimation.} 
		For each state-action pair  $(s,a)\in \cS\times \cA$, draw $N$ independent samples from the generative model: 
		$$
			s^{(i)}(s,a) \overset{\mathrm{i.i.d.}}{\sim} P(\cdot \mymid s,a), \qquad i=1,\ldots, N,
		$$ 
		and construct an empirical transition kernel $\widehat{P}$ as follows
		\begin{equation}
			\forall s' \in \mathcal{S}, \qquad \widehat{P}(s^\prime \mymid s, a) = \frac{1}{N}\sum_{i=1}^N \ind \big\{  s^{\prime} = s^{(i)}(s,a) \big\}.
			\label{eq:defn-empirical-P}
		\end{equation}
		In words, $\widehat{P}(s^\prime \mymid s,a)$ records the empirical frequency of transitions from $(s,a)$ to state $s^\prime$.  
		This leads to an empirical MDP $\widehat{\mathcal{M}} = (\cS,\cA, \widehat{P}, r,\gamma)$ constructed from the data samples. 
		We shall denote by $\widehat{V}^{\pi}$ the value function of any policy $\pi$ in the MDP $\widehat{\mathcal{M}}$.

	\item[2)] {\em Planning.} Compute the optimal policy
\begin{align}
	\widehat{\pi}\coloneqq \arg\max_\pi \; \widehat{{V}}^{\pi}
	\label{defn:pi-p-star-noperturb}
\end{align}
 of the empirical MDP $\widehat{\mathcal{M}}$, using, e.g., QVI or PI. 
	
\end{itemize}
See Algorithm~\ref{alg:policy-learn-model-based} for the full procedure.

\begin{algorithm}[t]
	\begin{algorithmic}[1] 
	\STATE\textbf{input:} discount factor $\gamma$, reward function $r$, sample size per state-action pair $N$. 
		\FOR{$(s,a) \in \cS \times \cA$}
		\STATE{ draw $N$ independent samples $s^{(i)}(s,a) \overset{\mathrm{i.i.d.}}{\sim} P(\cdot \mymid s, a)$. }
		\STATE{ compute $\widehat{P}(\cdot \mymid s,a)$ based on \eqref{eq:defn-empirical-P}.}
	\ENDFOR

	\STATE compute $\widehat{\pi}\coloneqq \arg\max_\pi \widehat{{V}}^{\pi}$ of the MDP $\widehat{\mathcal{M}} = (\cS,\cA, \widehat{P}, r,\gamma)$, 
		where $\widehat{{V}}^{\pi}$ represents the value function of policy $\pi$ in $\widehat{\mathcal{M}}$.  

   \end{algorithmic} 
    \caption{Model-based policy learning}
 \label{alg:policy-learn-model-based}
\end{algorithm}

We shall also present a ``randomly perturbed'' variant of the above model-based algorithm, where random reward perturbation is employed to facilitate analysis.  More specifically, when constructing the empirical MDP, we randomly perturb the immediate reward for each state-action pair $(s,a)$ as follows: 
		\begin{equation}
			r_{\mathrm{p}} (s,a) = r(s,a) + \zeta(s,a), \qquad \zeta(s,a) \overset{\mathrm{i.i.d.}}{\sim} \mathsf{Unif}(0,\xi),
			\label{eq:perturbed-reward-gone}
		\end{equation}
		where $\mathsf{Unif}(0,\xi)$ denotes the uniform distribution between 0 and some parameter $\xi>0$ (to be specified momentarily).\footnote{{Note that perturbation is only invoked when running the planning algorithms
and does not require collecting new samples. }} 
The perturbed empirical MDP is taken to be
$\widehat{\mathcal{M}}_{\mathrm{p}}=(\cS,\cA, \widehat{P}, {r}_{\mathrm{p}}, \gamma)$, using the probability transition kernel $\widehat{P}$ (cf.~\eqref{eq:defn-empirical-P}) and the perturbed reward function ${r}_{\mathrm{p}}=[{r}_{\mathrm{p}}]_{(s,a)\in \cS\times \cA}$. 
The perturbed model-based approach then returns the optimal policy
\begin{align}
	\widehat{\pi}_{\mathrm{p}}^{\star} \coloneqq \arg\max_\pi \; \widehat{{V}}^{\pi}_{{\mathrm{p}}}
	\label{defn:pi-p-star-perturb}
\end{align}
of the perturbed empirical MDP $\widehat{\mathcal{M}}_{\mathrm{p}}$, 
where  $\widehat{V}^{\pi}_{\mathrm{p}}$  denotes the value function of any policy $\pi$ in the MDP $\widehat{\mathcal{M}}_{\mathrm{p}}$. 
The whole procedure is summarized in Algorithm~\ref{alg:policy-learn-model-based-perturbed}. 
It is self-evident that this variant has the same sample size as the unperturbed counterpart described above.

\begin{algorithm}[t]
	\begin{algorithmic}[1] 
	\STATE\textbf{input:} discount factor $\gamma$, reward function $r$, sample size per state-action pair $N$. 
		\FOR{$(s,a) \in \cS \times \cA$}
		\STATE{ draw $N$ independent samples $s^{(i)}(s,a) \overset{\mathrm{i.i.d.}}{\sim}  P(\cdot \mymid s, a)$. }
		\STATE{ compute $\widehat{P}(\cdot \mymid s,a)$ based on \eqref{eq:defn-empirical-P}.}
		\STATE{ generate a perturbed immediate reward $r_{\mathrm{p}} (s,a)$ as \eqref{eq:perturbed-reward-gone}. } 
	\ENDFOR

	\STATE compute $\widehat{\pi}\coloneqq \arg\max_\pi \widehat{{V}}^{\pi}$ of the MDP $\widehat{\mathcal{M}}_{\mathrm{p}} = (\cS,\cA, \widehat{P}, r_{\mathrm{p}},\gamma)$, 
		where $\widehat{{V}}^{\pi}_{\mathrm{p}}$ represents the value function of policy $\pi$ in $\widehat{\mathcal{M}}_{\mathrm{p}}$.  

   \end{algorithmic} 
    \caption{Perturbed model-based policy learning}
 \label{alg:policy-learn-model-based-perturbed}
\end{algorithm}

\subsubsection{Theoretical guarantees} 
As it turns out, the class of model-based algorithms achieves minimal sample complexity. 
In fact, \citet{azar2013minimax} proved the first result demonstrating the asymptotic optimality of the unperturbed model-based algorithm (see Algorithm~\ref{alg:policy-learn-model-based}) for very small target precision $\varepsilon$. The $\varepsilon$-range of model-based algorithms was subsequently enlarged by \citet{sidford2018near,agarwal2020model}.  Later on, \citet{li2024breaking} developed the first theory confirming the full-range optimality of model-based algorithms by analyzing the perturbed variant in Algorithm~\ref{alg:policy-learn-model-based-perturbed}. This is presented in the theorem below, whose optimality shall be discussed momentarily.
\begin{theorem}[\cite{li2020breaking}]
\label{Thm:sample-compl-main}
	There exist some universal constants $c_0,c_1>0$ such that: for any $\delta > 0$ and any $0<\varepsilon \leq \frac{1}{1-\gamma}$, the policy $\widehat{\pi}_{\mathrm{p}}^{\star}$ defined in \eqref{defn:pi-p-star-perturb} obeys
	\begin{align}
		\forall (s,a) \in \mathcal{S} \times \mathcal{A}, \qquad 
		V^{\pihatstar_{\mathrm{p}}}(s)  &\geq  V^\star(s) - \varepsilon 
	\end{align}
	with probability at least $1-\delta$, provided that the perturbation size is $\xi = \frac{c_1(1-\gamma)\varepsilon}{S^5A^5}$ and that 
	the sample size per state-action pair exceeds 
	\begin{align}
	\label{EqnSamples-main}
		N \geq \frac{c_{0}}{(1-\gamma)^{3}\varepsilon^{2}}  \log\left(\frac{SA}{(1-\gamma)\varepsilon \delta}\right). 
	\end{align}
	In addition, both the QVI and PI algorithms w.r.t.~$\widehat{\mathcal{M}}_{\mathrm{p}}$ are able to recover $\widehat{\pi}_{\mathrm{p}}^{\star}$ perfectly within $O\big(\frac{1}{1-\gamma}\log(\frac{SA}{(1-\gamma)\varepsilon\delta}) \big)$ iterations.  
\end{theorem}

In a nutshell, the above theorem asserts that model-based algorithms succeed in finding an $\varepsilon$-optimal policy  
as soon as the total sample complexity exceeds the order of $\frac{SA}{(1-\gamma)^{3}\varepsilon^2}$ (modulo some log factor). It is worth emphasizing that, compared to prior literature, this result imposes no restriction on the range of $\varepsilon$ and, in particular, this allows the accuracy level $\varepsilon$ to go all the way up to $\frac{1}{1-\gamma}$. 
This theory is particularly useful in the regime with small-to-moderate sample sizes, since its validity is guaranteed as long as 
\begin{align}
	N \geq \widetilde{\Omega}\Big(\frac{1}{1-\gamma}\Big) . 
	\label{eq:sample-size-range-planning}
\end{align}

In fact, the model-based approach is provably minimax-optimal (up to some logarithmic factor) for the full $\varepsilon$-range, 
as confirmed by the following minimax lower bound originally developed by \citet{azar2013minimax}. 
%
%
\begin{theorem}[\cite{azar2013minimax}]
\label{theorem:azar}
	For any $0<\varepsilon<\frac{1}{1-\gamma}$, $0<\delta<1$ and any $(S,A,\gamma)$, there exists some MDP such that no algorithm can find a policy $\pi^{\mathsf{est}}$ achieving $\|V^{\pi^{\mathsf{est}}} - V^{\star}\|_\infty \leq \varepsilon$ with probability at least $1-\delta$, unless the total number of samples exceeds
\begin{align}
\label{eqn:minimax-lb}
	\frac{SA}{c_1 (1-\gamma)^3 \varepsilon^2} \log \left(\frac{SA}{c_2 \delta} \right) .
\end{align}
	Here, $c_1, c_2>0$ are some universal constants.
\end{theorem}



\subsection{The model-free approach}
\label{chapter:generative-model-MF}

We now turn to the model-free approach.  In stark contrast to the model-based approach, 
 model-free algorithms sidestep the model estimation stage and proceed to estimate the value function and Q-function directly. 
We shall concentrate on 
Q-learning \citep{watkins1989learning,watkins1992q}, which is arguably one of the most widely adopted model-free algorithms.

\subsubsection{Algorithm: Q-learning}

In a nutshell, Q-learning applies stochastic approximation techniques \citep{robbins1951stochastic} to find the fixed point of the Bellman operator. 
It is an iterative algorithm that maintains a Q-function estimate $Q_t: \cS\times \cA \rightarrow \mathbb{R}$ for all $t\geq 0$. 
Given access to a generative model, the Q-learning algorithm adopts the following update rule: 
in each iteration $t$, the algorithm draws an independent sample 
$$s_t(s,a) \sim P(\cdot \mymid s,a)$$ 
for every state-action pair $(s,a)\in \cS\times \cA$, and updates  
{\em all} entries of the Q-function estimate at once via the following update rule
\begin{equation}
\label{eqn:q-learning}
	Q_t  = (1- \eta_t ) Q_{t-1} + \eta_t \mathcal{T}_t (Q_{t-1}) . 
\end{equation}
Here, $\eta_t \in (0,1]$ denotes the learning rate or the step size in the $t$-th iteration,  
and $\mathcal{T}_t$ denotes the empirical Bellman operator constructed by samples collected in the $t$-th iteration, i.e.,
\begin{align}  \label{defn:empirical-Bellman-t-inf}
	\mathcal{T}_t (Q) (s , a ) &\coloneqq 
		 r(s , a ) + \gamma \max_{a' \in \cA} Q\big(s_t(s,a), a'\big) 
\end{align}
for each state-action pair $(s,a)\in \cS\times \cA$.
In words, $\mathcal{T}_t(Q)$ forms an empirical estimate of the Bellman operator \eqref{eq:bellman_optimality_operator} applied to $Q$.  
		 The corresponding value function estimate $V_t: \cS \rightarrow \mathbb{R}$ in the  $t$-th iteration is defined as
\begin{align}
	\label{defn:Vt}
	\forall s\in \cS: \qquad V_t(s) :=  \max_{a\in \cA} Q_{t}(s,a) .
\end{align}
%
		 
%
In addition, the algorithm is initialized in a way that obeys $0\leq Q_0(s,a) \leq \frac{1}{1-\gamma}$ for every state-action pair $(s,a)$. The complete description of Q-learning is summarized in Algorithm~\ref{alg:q-infinite}. 
Note that this version of Q-learning is frequently referred to as {\em synchronous} Q-learning, since all entries of the Q-function are updated simultaneously.
 
\begin{remark}
	Another strand of work studied {\em asynchronous} Q-learning and TD learning, which concern the case where the samples take the form of a Markovian trajectory. The interested reader is referred to \cite{tsitsiklis1994asynchronous,tsitsiklis1997analysis,beck2012error,bhandari2021finite,srikant2019finite,qu2020finite,li2021sample,even2003learning,chen2021lyapunov,mou2020linear,yan2023efficacy,wu2024statistical,wu2025uncertainty,li2024q,chen2025non}, with the state-of-the-art sample complexity derived in \citet{li2024q}. 
\end{remark}

\begin{algorithm}[t]
	\begin{algorithmic}[1] 
	\STATE\textbf{inputs:} learning rates $\{\eta_t\}$, number of iterations $T$, initialization $Q_0$.
   \FOR{$t=1,2,\cdots,T$}
	\STATE{ Draw  $s_t(s,a) \sim  P(\cdot \mymid s,a)$ for each $(s,a)\in \cS\times \cA$. }
	\STATE{ Compute $Q_t$ according to \eqref{eqn:q-learning} and \eqref{defn:empirical-Bellman-t-inf}.}
\ENDFOR
   \end{algorithmic} 
    \caption{Synchronous Q-learning}
 \label{alg:q-infinite}
\end{algorithm}

\subsubsection{Theoretical guarantees ($A\geq 2$)}

Given the popularity of Q-learning, extensive theoretical analyses have been carried out to characterize the performance of synchronous Q-learning (e.g., \cite{even2003learning,beck2012error,wainwright2019stochastic,chen2020finite,li2024q}). In what follows, we present the state-of-the-art  $\ell_{\infty}$-based sample complexity bound for synchronous Q-learning. 
\begin{theorem}[\cite{li2024q}]
\label{thm:infinite-horizon-simple}
Consider any $\delta\in(0,1)$,  $\varepsilon\in(0,1]$, and $\gamma\in [1/2,1)$. Suppose that $A\geq 2$, and that for any $0\leq t\leq T$, the
learning rates satisfy
\begin{subequations}
\label{eq:thm-infinite-horizon-condition}
\begin{equation}
\frac{1}{1+\frac{c_1 (1-\gamma) T}{\log^3 T}}\le\eta_{t}\le\frac{1}{1+\frac{c_2 (1-\gamma) t}{\log^3 T}} \label{eq:thm:infinite-horizaon-eta}
\end{equation}
for some small enough universal constants $c_{1}\geq c_{2}>0$. 
Assume that the total number of iterations $T$ obeys
\begin{equation}
	T\ge\frac{c_{3}\big(\log^{4}T\big)\big(\log\frac{SAT}{\delta}\big)}{(1-\gamma)^{4}\varepsilon^{2}}
	\label{eq:thm:infinite-horizon-T}
\end{equation}
for some sufficiently large universal constant $c_{3}>0$. 
\end{subequations}
If the initialization obeys $0\leq {Q}_{0}(s,a) \leq\frac{1}{1-\gamma}$ for all $(s,a)\in \cS\times \cA$, 
then Algorithm \ref{alg:q-infinite} achieves
$ \| {Q}_{T} -  {Q}^{\star} \|_{\infty} \le\varepsilon$ with probability at least $1-\delta$. 
\end{theorem}

In a nutshell, Theorem~\ref{thm:infinite-horizon-simple} develops a non-asymptotic bound on the iteration complexity of Q-learning in the presence of the generative model.  
Recognizing that $SA$ independent samples are drawn in each iteration, 
we can see from Theorem~\ref{thm:infinite-horizon-simple} the following sample complexity bound
\begin{equation}
	\widetilde{O}\Big( \frac{SA}{(1-\gamma)^4 \varepsilon^2} \Big)
	\label{eq:sample-complexity-sync-Q-learning}
\end{equation}
is sufficient for Q-learning to attain $\varepsilon$-accuracy ($0<\varepsilon<1$) in an entrywise sense.   

We shall also take a moment to discuss the learning rate schedule studied above. Theorem~\ref{thm:infinite-horizon-simple} accommodates two commonly adopted learning rate schemes (cf.~\eqref{eq:thm:infinite-horizaon-eta}): (i) linearly rescaled learning rates $\frac{1}{1+\frac{c_2 (1-\gamma)}{\log^2 T}t}$, and (ii) iteration-invariant learning rates $\frac{1}{1+\frac{c_1 (1-\gamma) T}{\log^2 T}}$ (which depend on the total number of iterations $T$ but not the iteration number $t$).
In particular, when $T=\frac{c_{3} (\log^{4}T)\big(\log\frac{SAT}{\delta}\big)}{(1-\gamma)^{4}\varepsilon^{2}}$, the constant learning rates can be taken to be on the order of
\[
	\eta_t \equiv \widetilde{O} \big( (1-\gamma)^3 \varepsilon^2 \big),  \qquad 0\leq t\leq T, 
\]
which depends almost solely on the discount factor $\gamma$ and the target accuracy $\varepsilon$. 
Interestingly, both learning rate schedules lead to the same $\ell_{\infty}$-based sample complexity bound (in an order-wise sense), making them appealing for practical use.

\subsubsection{A matching algorithmic lower bound and sub-optimality ($A\geq 2$)} 
The careful reader might remark that there remains a gap between the sample complexity bound for Q-learning (cf.~Theorem~\ref{thm:infinite-horizon-simple})  and the minimax lower bound (cf.~Theorem~\ref{theorem:azar}). More specifically, the minimax lower bound scales on the order of $\frac{SA}{(1-\gamma)^3\varepsilon^2}$ and is achievable---up to some logarithmic factor---by the model-based approach \citep{azar2013minimax,agarwal2019optimality,li2020breaking}. 
This raises natural questions regarding whether the sample complexity bound of Q-learning can be further improved, and whether there is any intrinsic bottleneck that prevents vanilla Q-learning from attaining optimal performance.  
To answer these questions, we present the following lower bound for plain Q-learning developed by \citet{li2024q}, 
with the aim of confirming the sharpness of Theorem~\ref{thm:infinite-horizon-simple} and revealing the sub-optimality of Q-learning. 
\begin{theorem}[\cite{li2024q}]
\label{thm:LB-example}
	Assume that $3/4\leq \gamma < 1$ and that $T\geq \frac{c_3}{(1-\gamma)^2}$ for some sufficiently large constant $c_3>0$. 
	Suppose that the initialization is $Q_0 \equiv 0$, and that the learning rates are taken to be either (i) $\eta_t = \frac{1}{1+\ceta (1-\gamma)t}$ for all $t\geq 0$, or (ii) $\eta_t \equiv \eta$ for all $t\geq 0$. 
	There exists a $\gamma$-discounted MDP with $S=4$ and $A=2$ such that Algorithm \ref{alg:q-infinite}---with any $\ceta >0$ and any $\eta \in (0,1)$---obeys
\begin{equation}
	\max_{s\in \cS} \mathbb{E} \Big[ \big|V_T(s) - V^{\star}(s)\big|^2 \Big] 
	\geq  \frac{c_{\mathsf{lb}}}{(1-\gamma)^4T\log^2 T} ,
	\label{eq:lower-bounding-VT-thm-sync}
\end{equation}
where $c_{\mathsf{lb}} > 0$ is some universal constant.
\end{theorem}
Theorem~\ref{thm:LB-example} provides an {\em algorithm-dependent} lower bound for vanilla Q-learning. 
As asserted by this theorem, it is impossible for Q-learning to attain $\varepsilon$-accuracy (in the sense that $\max_s \mathbb{E} \big[ \big|V_T(s) - V^{\star}(s)\big|^2 \big] \leq \varepsilon^2$) unless the number of iterations exceeds the order of 
\[
	\frac{1}{(1-\gamma)^{4}\varepsilon^{2}}
\]
up to some logarithmic factor. Consequently, the performance guarantees for Q-learning derived in Theorem~\ref{thm:infinite-horizon-simple} are sharp in terms of the dependency on the effective horizon $\frac{1}{1-\gamma}$.  
On the other hand, this reveals the sub-optimality of plain Q-learning, 
whose horizon scaling is above that of the minimax lower bound by a multiplicative factor of $\frac{1}{1-\gamma}$. 
Hence, more sophisticated algorithmic tricks are necessary to further reduce the sample complexity. 
For instance, a variance-reduced variant of Q-learning---namely, leveraging the idea of variance reduction originating from stochastic optimization \citep{johnson2013accelerating} to accelerate convergence---has been shown to attain minimax optimality 
\eqref{eqn:minimax-lb} for any $\varepsilon \in (0,1]$; see \citet{wainwright2019variance} for more details. 
We note that the sample inefficiency of Q-learning has been long noted \citep{thrun1993issues,hasselt2010double}, due to the over-estimation bias when replacing the expectation in $\max_{a\in\cA} \mathbb{E}[Q(s,a)]$ by its empirical estimates; see, for example, \citet{hasselt2010double} for more discussion.


\subsubsection{A special case: temporal difference learning ($A=1$)}
The careful readers might already note that the lower bound in Theorem~\ref{thm:LB-example} does not apply to the case with $A=1$. 
In fact, when the action set $\cA$ is a singleton, synchronous Q-learning reduces to the celebrated temporal difference (TD) learning algorithm designed for value estimation \citep{sutton1988learning,tsitsiklis1997analysis,bhandari2021finite}. 
Note that when $A=1$, the MDP reduces to a Markov reward process (MRP) \citep{bertsekas2017dynamic}, and we shall abuse the notation to use  
 $P: \mathcal{S} \rightarrow \Delta(\cS)$ to describe the probability transition kernel, and employ $r: \mathcal{S} \rightarrow [0,1]$ to represent the reward function (with $r(s)$ indicating the immediate reward gained in state $s$). 
 TD learning maintains an estimate $V_t: \mathcal{S} \rightarrow \mathbb{R}$ of the value function in each iteration $t$,\footnote{There is no need to maintain additional Q-estimates, as the Q-function and the value function coincide when $A=1$. } 
and carries out the following iterative update rule
\begin{align}
\label{eqn:td-learning}
	V_t (s) &= (1- \eta_t ) V_{t-1} (s) + \eta_t \big( r(s) + \gamma   V_{t-1}\big(s_t(s)\big) \big)
\end{align}
for each state $s\in \cS$. 
As before, $\eta_t\in (0,1]$ is the learning rate at time $t$, the initial estimate $V_0(s)$ is taken to be within $\big[0, \frac{1}{1-\gamma}\big]$, and in each iteration, 
the samples $\{s_t(s) \mymid s\in \cS\}$ are generated independently.  
 The whole algorithm of TD learning is summarized in Algorithm~\ref{alg:td-infinite}.

 \begin{algorithm}[t]
	\begin{algorithmic}[1] 
	\STATE\textbf{inputs:} learning rates $\{\eta_t\}$, number of iterations $T$,  initialization $V_0$. 
   \FOR{$t=1,2,\cdots,T$}
	\STATE{ Draw  $s_t(s) \sim  P(\cdot \mymid s)$ for each $s\in \cS$. }
	\STATE{ Compute $V_t$ according to \eqref{eqn:td-learning}.}
\ENDFOR
   \end{algorithmic} 
    \caption{Synchronous TD learning.}
 \label{alg:td-infinite}
\end{algorithm}

We now present the $\ell_{\infty}$-based sample complexity of synchronous TD learning.

\begin{theorem}[\cite{li2024q}]
\label{thm:policy-evaluation}
	Consider any $\delta\in(0,1)$,  $\varepsilon\in(0,1]$, and $\gamma\in [1/2,1)$. Suppose that for any $0\leq t\leq T$, the
learning rates satisfy
\begin{subequations}
\label{eq:thm-td}
\begin{equation}
\frac{1}{1+\frac{c_1 (1-\gamma) T}{\log^2 T}}\le\eta_{t}\le\frac{1}{1+\frac{c_2 (1-\gamma) t}{\log^2 T}} \label{eq:thm:eta-TD}
\end{equation}
for some small enough universal constants $c_{1}\geq c_{2}>0$. 
Assume that the total number of iterations $T$ obeys
\begin{equation}
	T\ge\frac{c_{3}\big(\log^{3}T\big)\big(\log\frac{S T}{\delta}\big)}{(1-\gamma)^{3}\varepsilon^{2}}
	\label{eq:thm:sample-size-TD-thm}
\end{equation}
for some sufficiently large universal constant $c_{3}>0$. 
\end{subequations}
If the initialization obeys $0\leq {V}_{0}(s) \leq\frac{1}{1-\gamma}$ for all $s\in \cS$, 
then  with probability at least $1-\delta$, Algorithm~\ref{alg:td-infinite} achieves
$\| {V}_{T}  -  {V}^{\star} \| \le\varepsilon$.

\end{theorem}

Given that each iteration of synchronous TD learning makes use of $S$ samples, 
Theorem~\ref{thm:policy-evaluation} implies that the sample complexity of TD learning is at most
\begin{equation}
	\label{eq:TD-sample-complexity-simple}
	\widetilde{O} \bigg( \frac{S}{(1-\gamma)^3 \varepsilon^2} \bigg)
\end{equation}
for any target accuracy level $\varepsilon \in (0,1]$. 
This non-asymptotic result is valid as long as
the learning rates are chosen to be either a proper constant or rescaled linear (see~\eqref{eq:thm:eta-TD}). Compared with the minimax lower bound 
\begin{equation}
	\label{eq:TD-sample-complexity-simple-LB}
	\widetilde{\Omega} \bigg( \frac{S}{(1-\gamma)^3 \varepsilon^2} \bigg),
\end{equation}
somewhat surprisingly, Theorem~\ref{thm:policy-evaluation} unveils the minimax optimality of the sample complexity (modulo some logarithmic factor) of TD learning  for the synchronous setting, 
which stands in stark contrast to the sub-optimality of Q-learning when $A\geq 2$.

\section{Online reinforcement learning}
\label{chapter:online-RL}

We now turn attention to the online episodic setting. 
In contrast to the simulator setting that permits sampling of any state-action pair,  
an agent in online episodic RL is only allowed to draw sample trajectories by executing a policy in the unknown MDP. 
Careful deliberation needs to be undertaken when deciding what policies to use to allow for effective interaction with the unknown environment, and how to optimally balance exploitation and exploration under limited sampling budget.

\subsection{Problem formulation}

In this section, we focus on online tabular RL in finite-horizon nonstationary MDPs, a setting that has been solved even in the most sample-limited regime. Its counterpart in discounted infinite-horizon MDPs has not been fully settled (see, e.g., \cite{lattimore2012pac,strehl2006pac,dong2019q,zhang2021model,szita2010model,ji2023regret2}) and calls for further investigation .

To be concrete, consider the online episodic RL setting formulated in a finite-horizon nonstationary MDP $\mathcal{M}=(\mathcal{S},\mathcal{A},H, \{P_h\}_{h=1}^H, \{r_h\}_{h=1}^H)$, where the agent is allowed to execute the MDP sequentially in a total number of $K$ episodes each of length $H$. 
This amounts to collecting  $$T=KH\text{ samples}$$ in total. 
More specifically, in each episode $k=1,\ldots, K$, the agent is assigned an independent initial state $s_1^k \sim \mu$, 
and selects a policy $\pi^k =\{\pi_h^k \}_{h=1}^H$  
learned based on the information collected up to the $(k-1)$-th episode. 
The $k$-th episode is then executed following the policy $\pi^k$ and the dynamic of the MDP $\mathcal{M}$, 
leading to a length-$H$ sample trajectory $\big\{ (s_h^k,a_h^k,r_h^k) \big\}_{1\leq h\leq H}$, 
with $s_h^k$, $a_h^k$ and $r_h^k=r_h(s_h^k,a_h^k)$ denoting the state, action and immediate reward at step $h$ of this episode.

To evaluate the learning performance, 
a widely used metric is the (cumulative) regret over all $K$ episodes:   
\begin{align}
	\mathsf{Regret}(K) \coloneqq \sum_{k=1}^K \left( V^{\star}_{1}(s_1^k) -V^{\pi^k}_1(s_1^k) \right),
	\label{eq:defn-regret}
\end{align}
and our goal is to design an online RL algorithm that minimizes $\mathsf{Regret}(K)$ regardless of the allowable sample size $K$.




\subsection{Minimax lower bound}
To provide a theoretical baseline,
we first present a minimax regret lower bound \citep{jin2018q,domingues2021episodic}:
\begin{align}
	\text{(minimax lower bound)} \qquad 
	\Omega\left(\min\big\{\sqrt{SAH^3K} ,\, HK \big\}\right), 
	\label{eq:minimax-lower-bound-1}
\end{align}
assuming that the immediate reward at each step falls within $[0,1]$ and imposing no restriction on $K$.  
Given that a regret of $O(HK)$ can be trivially achieved (as the sum of rewards in any $K$ episodes cannot exceed $HK$), 
we shall sometimes drop the $HK$ term and simply write
\begin{align}
	\text{(minimax lower bound)} \qquad 
	\Omega\big(\sqrt{SAH^3K} \big) \qquad \text{if }K \geq SAH. 
	\label{eq:minimax-lower-bound-2}
\end{align}
%




\subsection{Model-based online RL}


\paragraph{The optimism principle in the face of uncertainty.} 
In order to ensure a sufficient degree of exploration, a popular idea is 
the optimism principle in the face of uncertainty, which has received widespread adoption  from bandits to online RL  \citep{lai1985asymptotically,lattimore2020bandit,agarwal2019reinforcement}. 
In a nutshell, the optimism principle advocates action selection based on the so-called ``optimistic'' Q-function estimate --- which is an upper confidence bound on the Q-function  --- rather than the Q-function estimate itself. The incorporation of the confidence interval helps encourage exploration of state-action pairs that have been insufficiently explored, offering a simple strategy to balance  exploitation with exploration. In what follows, we shall briefly describe how the optimism principle can be incorporated into the model-based approach.

\paragraph{A first attempt: UCBVI.}
%
%
In an attempt to leverage the optimism principle in model-based algorithms, 
\citet{azar2017minimax} proposed a model-based algorithm called upper confidence bound value iteration (UCBVI), 
which was the first online RL algorithm to achieve asymptotically optimal regret. 
On a high level, this algorithm implements the optimism principle in the face of uncertainty 
		by adopting the frequently used UCB-based framework. 
More specifically, in each iteration, the learner performs the following: 
\begin{itemize}
	\item {\em Model estimation:} Compute an empirical estimate $\widehat{P}=\{\widehat{P}_h\in \mathbb{R}^{SA\times S}\}_{1\leq h\leq H}$ of the transition probability kernel using all samples that have been collected; 

	\item {\em Opportunistic Q-function estimates:} 
		Maintain upper estimates for the associated value function and Q-function using  
\begin{subequations}
\label{eq:Qh-Vh-UCB-informal}	
\begin{align}
Q_{h}(s,a)\, & := \,\min\big\{{r}_{h}(s,a)+\langle\widehat{P}_{s,a,h},V_{h+1}\rangle+b_{h}(s,a),H\big\}, 
	\label{eq:Qh-UCB-informal}\\
V_{h}(s)\, & :=\, \max\nolimits_{a}Q_{h}(s,a) 
\end{align}
\end{subequations}
for all state-action pairs. 
Here, $Q_{h}$ (resp.~$V_h$) indicates the running estimate for the Q-function (resp.~value function), 
		whereas $b_{h}(s,a)\geq 0$ is some suitably chosen bonus term (typically based on the Berstein inequality) that compensates for the uncertainty.  

	\item {\em Updates of policy estimates, and sampling:} 
		The above opportunistic Q-estimate in turn allows one to obtain a new policy estimate (via a simple greedy rule), 
which is then exeuted to collect a new episode of data samples.

\end{itemize}


\noindent 
Noteworthily, the empirical model $\widehat{P}$ shall be updated many times as new samples continue to arrive, 
and hence the updating rule \eqref{eq:Qh-Vh-UCB-informal} will be invoked often as well.

As alluded to previously, this UCBVI algorithm provably achieves asymptotically optimal regret; more precisely, it enjoys a regret bound $\widetilde{O}\big(\sqrt{SAH^3K} + H^3S^2A\big)$. 
A close inspection reveals that this regret matches the minimax lower bound \eqref{eq:minimax-lower-bound-1} if and only if 
\begin{align}
	\big(\text{burn-in cost of \citet{azar2017minimax}}\big) \qquad K \gtrsim S^3A H^3, 
\end{align}
due to the presence of the lower-order term $H^3S^2A$ in the regret bound. 
This burn-in cost is clearly undesirable, 
since the sample size available in many practical scenarios might be far below this requirement.

\paragraph{A regret-optimal model-based algorithm without burn-ins: MVP.} 
The huge theoretical burn-in cost of UCBVI has motivated a number of follow-ups to pursue theoretical improvements. 
Encouragingly, \citet{zhang2024settling,zhang2025settling} proposed an opportunistic model-based algorithm---called {MVP}: {\em Monotonic Value Propagation} \citep{zhang2020reinforcement}---that achieve provably minimax-optimal regret (up to some logarithmic factor) regardless of the number $K$ of episodes that can be collected.  
The full procedure can be found in \citet{zhang2025settling}. Due to space limitations, we shall only point out several key ingredients of MVP that differ from UCBVI.

\begin{itemize}

	\item {\em An epoch-based procedure and a doubling trick.} 
		Compared to the original UCBVI in \citep{azar2017minimax}, 
		one distinguishing feature of MVP is to update the empirical transition kernel and empirical rewards in an epoch-based fashion, 
		as motivated by a doubling update framework adopted in \citet{jaksch2010near}. 
		More concretely, the whole learning process is divided into consecutive epochs via a simple doubling rule; 
		namely, whenever there exits a $(s,a,h)$-tuple whose visitation count reaches a power of 2, MVP ends the current epoch,  reconstructs the empirical model, computes the Q-function and value function using the newly updated transition kernel and rewards, and then starts a new epoch with an updated sampling policy. 
		This stands in stark contrast with the original UCBVI, which computes new estimates for the transition model, Q-function and value function in every episode. With this doubling rule in place, the estimated transition probability vector for each $(s,a,h)$-tuple will be updated by no more than $\log_2K$ times, 
		a feature that plays a pivotal role in significantly reducing some sort of covering number needed in the covering-based analysis. In each epoch, the learned policy is induced by the optimistic Q-function estimate---computed based on the empirical transition kernel of the {\em current} epoch---which will then be employed to collect samples in {\em all} episodes of the next epoch.

	\item {\em Monotonic bonus functions.} Another crucial step in order to ensure near-optimal regret lies in careful designs of the data-driven bonus terms $\{b_h(s,a)\}$. 
		Here, the MVP algorithm adopts the monotonic Bernstein-style bonus function, computed based on the Bernstein inequality in a data-driven manner. 
Note that in order to enable variance-aware regret, this algorithm also needs to keep track of the empirical variance of the (stochastic) immediate rewards. 

\end{itemize}


Let us take a moment to discuss several prior online RL algorithms based on the optimism principle. 
In the context of online RL, 
\citet{jaksch2010near} 
constructed confidence regions for the probability transition kernel to help select optimistic policies in the setting of weakly communicating MDPs, 
based on a variant  of the UCRL algorithm originally proposed in 
\citet{auer2006logarithmic}.
When applied to episodic finite-horizon MDPs, the regret bound in \citet{jaksch2010near} was suboptimal by a factor of at least $\sqrt{H^2S}$. 
In comparison, a more sample-efficient paradigm is to build Bernstein-style UCBs for the optimal values to help select exploration policies (more details can be found around Eq.~\eqref{def:bonus-Bernstein-infinite}), 
which has been recently adopted in both model-based \citep{azar2017minimax} and model-free algorithms \citep{jin2018q}. 
Note that Bernstein-style uncertainty estimation alone is not enough to ensure regret optimality in model-free algorithms, 
thereby motivating the design of more sophisticated variance reduction strategies \citep{zhang2020almost,li2021breaking}. 
	How to optimally leverage the optimism principle to achieve full-range optimal regret was an open problem prior to \citet{zhang2020reinforcement}. The use of the optimism principle has further been studied in the context of reward-free RL and the case with linear function approximation; see, e.g., \cite{jin2020reward,zhang2020task,li2024minimax,jin2020provably,he2023nearly,wang2020reward,zhang2021reward,yang2020reinforcement} and the references therein. 


\subsection{Theoretical guarantees}

Encouragingly, the model-based algorithm MVP described above  
attains the minimax-optimal regret regardless of the number $K$ of episodes that can be collected. 
%
\begin{theorem}[\cite{zhang2024settling,zhang2025settling}]\label{thm1}  
For any $K \ge 1$ and any $0<\delta<1$, 
	there exists an algorithm (see the MVP algorithm in \citet{zhang2024settling}) obeying
\begin{equation}
	\mathsf{Regret}(K) 
	\lesssim \min\bigg\{\sqrt{SAH^{3}K \log^5\frac{SAHK}{\delta}},HK\bigg\}
	\label{eq:regret-Thm1}
\end{equation}
with probability at least $1-\delta$.
\end{theorem}
The optimality of the regret bound \eqref{eq:regret-Thm1} can be readily seen given that 
it matches the minimax lower bound \eqref{eq:minimax-lower-bound-1} (modulo some logarithmic factor). 
One can also easily translate the above regret bound into sample complexity or PAC (which stands for ``probably approximately correct'') bounds, which count the number of episodes needed to find an $\varepsilon$-optimal policy $\widehat{\pi}$ in the sense that $\mathbb{E}_{s_1 \sim \mu}\big[V_1^{\star}(s_1) - V^{\widehat{\pi}}_1(s_1)\big] \le \varepsilon$. The MVP algorithm is able to return an $\varepsilon$-suboptimal policy with high probability using at most 
\begin{equation}
	\text{(sample complexity)}\qquad 
	\widetilde{O}\left(\frac{SAH^3}{\varepsilon^2}\right) \quad \text{episodes}
\end{equation}
(or equivalently, $\widetilde{O}\big(\frac{SAH^4}{\varepsilon^2}\big)$ sample transitions as each episode has length $H$).  
Remarkably, this result holds true for the entire $\varepsilon$ range (i.e., any $\varepsilon \in (0,H]$), essentially eliminating the need of any burn-in cost.  
It is noteworthy that even in the presence of an idealistic generative model (cf.~Section~\ref{sec:model-based-policy-eval-sim}), this order of sample size is un-improvable \citep{azar2013minimax,li2020breaking}.



\section{Offline reinforcement learning}
\label{chapter:offline-RL}

Despite the promise of online RL, limited capability of online data collection in other real-world applications presents a fundamental bottleneck for carrying such RL success over to broader scenarios. 
To circumvent this bottleneck, one plausible strategy is  {\em offline RL} or {\em batch RL} \citep{lange2012batch,levine2020offline}, which makes more effective use of data collected previously,
given that such historical data might contain useful information that readily transfers to new tasks (for instance, the state transitions in a historical task might sometimes resemble what happen in new tasks).  
The potential of this data-driven approach has been explored and recognized in a diverse array of contexts including but not limited to robotic manipulation, autonomous driving, and healthcare; see \citet{levine2020offline,prudencio2022survey} for overviews of recent development.  
 A desired offline RL algorithm would be able to achieve the target statistical accuracy using as few samples as possible. 


\subsection{Problem formulation}

%

To set the stage and facilitate discussion, let us introduce one mathematical model for the infinite-horizon $\gamma$-discounted setting, and focus our analysis on this concrete model. 
%
%
Imagine that we observe a batch dataset $\mathcal{D} = \{(s_i,a_i,s_i')\}_{1\leq i\leq N}$ containing $N$ sample transitions. These samples are independently generated based on a distribution $\myrho \in \Delta(\cS\times \cA)$ and the transition kernel $P$ of the MDP, namely, 
		\begin{align}
			(s_i,a_i) \overset{\mathrm{ind.}}{\sim} \myrho 
			\qquad \text{and} \qquad
			s_i' \overset{\mathrm{ind.}}{\sim} P(\cdot \mymid s_i, a_i),
			\qquad  1\leq i\leq N.
			\label{eq:sampling-offline-inf-iid}
		\end{align}
		%
In addition, it is assumed that the learner is aware of the reward function, for simplicity.

\paragraph{Goal.} Armed with the batch dataset $\mathcal{D}$, the objective of offline RL in this case is to find a policy $\widehat{\pi}$ that attains near-optimal value functions---with respect to a given test state distribution $\rho\in \Delta(\cS)$---in a sample-efficient manner. 
To be precise, for a prescribed accuracy level $\varepsilon$, we seek to identify an $\varepsilon$-optimal policy $\widehat{\pi}$ satisfying, with high probability, 
\begin{align}
	V^\star(\rho) - V^{\widehat{\pi}}(\rho) \leq \varepsilon,
\end{align}
using a batch dataset $\mathcal{D}$ (cf.~\eqref{eq:sampling-offline-inf-iid}) containing as few samples as possible.
Particular emphasis is placed on achieving minimal sample complexity for the entire range of accuracy levels (namely, for any $\varepsilon \in \big(0,\frac{1}{1-\gamma}\big]$), so as to accommodate the sample-hungry regime. 


In contrast to online exploratory RL, 
offline RL has to deal with several critical issues resulting from the absence of active exploration. 
Below we single out two representative issues surrounding offline RL. 


\begin{itemize}

\item {\em Distribution shift.}
For the most part,  the historical data is generated by a certain behavior policy that departs from the target, optimal one. 
As a result, some state-action pairs that are essential under the optimal policy are insufficiently visited by the behavior policy. 
A key challenge in offline RL thus stems from the shift of data distributions:
how to leverage past data to the most effect, even though the distribution induced by the target policy differs from what we have available?

\item {\em Limited data coverage.} 
	Ideally, if the dataset contained sufficiently many data samples for every state-action pair, then there would be hope to simultaneously learn the performance of every policy.  Such a uniform coverage requirement, however, is oftentimes not only unrealistic (given that we can no longer change the past data) but also unnecessary (given that we might only be interested in identifying a single optimal policy).    

\end{itemize}

\paragraph{The pessimism principle in the face of uncertainty.} 
Whether one can effectively cope with distribution shift and insufficient data coverage becomes a major factor that governs the feasibility and statistical efficiency of offline RL.  
In order to address the aforementioned issues, 
a recent strand of theoretical work put forward the {\em principle of pessimism or conservatism} (originally proposed by \citet{jin2021pessimism,rashidinejad2021bridging}; see also \cite{xie2021policy,yin2021towards,shi2022pessimistic,yan2023efficacy,uehara2021pessimistic,yin2022near,yan2024model,jin2022policy}).  
This is reminiscent of the optimism principle in the face of uncertainty for online exploration \citep{lai1985asymptotically,jaksch2010near,azar2017minimax,bourel2020tightening,jin2018q,zhang2025settling}, but works for drastically different reasons.  
One plausible idea of the pessimism principle, which has been incorporated into both model-based and model-free approaches, is to penalize  value estimation of those state-action pairs that have been poorly covered. Informally speaking,  insufficient coverage of a state-action pair inevitably results in low confidence and high uncertainty in the associated value estimation, and it is hence advisable to act cautiously by tuning down the corresponding value estimate.  As will be seen in this section,
proper use of pessimism amid uncertainty brings several provable benefits:  
(i) it allows for a reduced sample size that adapts to the extent of distribution shift; 
(ii) as opposed to uniform data coverage, it only requires coverage of the part of the state-action space reachable by the target policy.

\subsection{A key metric: single-policy concentrability}
To capture the distribution shift between the desired occupancy measure and the data distribution, 
we introduce a key quantity previously introduced in \citet{rashidinejad2021bridging}. 
\begin{definition}[Single-policy concentrability] 
\label{assumption:concentrate-infinite-simple}
The single-policy concentrability coefficient of a batch dataset $\mathcal{D}$ is defined as
\begin{align}
	C^{\star} \coloneqq \max_{(s, a) \in \cS \times \cA}\, \frac{d^{\star}(s, a)}{\myrho(s, a)} ,
	\label{eq:concentrate-infinite-simple}
\end{align}
where $d^{\star}(s, a): = d_{\rho}^{\pi^\star}(s,a)$ is defined in \eqref{eq:visitation_dis_sa}.
Clearly, one necessarily has $C^{\star} \geq 1$. 
\end{definition}
In words, $C^{\star}$ measures the distribution mismatch in terms of the maximum density ratio over the entire state-action space. 
The batch dataset can be viewed as expert data when $C^{\star}$ approaches 1, meaning that the batch dataset is close to the target policy in terms of the induced distributions.   
Moreover, this coefficient $C^{\star}$ is referred to as the ``single-policy'' concentrability coefficient since it is concerned with a single policy $\pi^{\star}$; this is clearly a much weaker assumption compared to the all-policy concentrability assumption (as adopted in, e.g., \citet{munos2007performance,chen2019information}), the latter of which assumes a uniform density-ratio bound over all policies and requires the dataset to be highly exploratory.


%
%

\begin{remark}
 \citet{li2022settling} proposed an  improved version of $C^{\star}$ obtained by clipping, which is defined by $\Cstar_{\mathsf{clipped}} \coloneqq \max_{(s, a) \in \cS \times \cA}\frac{\min\big\{d^{\star}(s, a), \frac{1}{S}\big\}}{\myrho(s, a)}$ and satisfies $\Cstar_{\mathsf{clipped}} \leq \Cstar$. The theorems in Section~\ref{sec:VI-LCB-infinite-horizon} continue to hold with this refined notion. 
 
\end{remark}

\subsection{Model-based offline RL}
\label{sec:VI-LCB-infinite-horizon}

Both model-based and model-free algorithms have been recently developed to tackle offline RL. 
In this section, we introduce a simple model-based offline RL algorithm that incorporates lower concentration bounds in value estimation. 
This algorithm, which is dubbed as VI-LCB, applies value iteration (based on some pessimistic Bellman operator) to the empirical MDP, 
with the key ingredients described below.

\paragraph{The empirical MDP.} Recall that we are given $N$ independent sample transitions $\{(s_i, a_i, s_i')\}_{i = 1}^N$ in the dataset $\mathcal{D}$.
For any given state-action pair $(s,a)$, we denote by 
\begin{align}
N(s, a) \coloneqq \sum_{i = 1}^{N} \ind\big((s_i, a_i) = (s, a)\big)
	\label{eq:defn-Nsa-infinite}
\end{align}
the number of samples transitions from $(s, a)$.
We then construct an empirical transition matrix $\widehat{P}$ such that
\begin{align*}
\forall (s,a,s')\in \cS\times \cA \times \cS: \;	\widehat{P}(s'\mymid s,a) = 
	\begin{cases} \frac{1}{N(s,a)} \sum\limits_{i=1}^N \mathds{1} \big\{ (s_i, a_i, s_i') = (s,a,s') \big\}, & \text{if } N(s,a) > 0 \\
		\frac{1}{S}, & \text{else}
	\end{cases}.
\end{align*}
%

\paragraph{The pessimistic Bellman operator.} 
The offline algorithm presented here seeks to approximate the fixed point of 
some variant of the classical Bellman operator.   
Let us first introduce this key operator and elucidate how the pessimism principle is enforced.  
Recall that the Bellman operator  $\mathcal{T}(\cdot): \mathbb{R}^{SA}\rightarrow \mathbb{R}^{SA} $ w.r.t.~the transition kernel $P$ (cf.~\eqref{eq:bellman_optimality_operator}) is defined such that
for any vector $Q \in \mathbb{R}^{SA}$, 
\begin{align}
	\mathcal{T}(Q)(s, a) \coloneqq r(s, a) + \gamma P_{s, a}V \qquad\text{for all }(s, a)\in\cS\times\cA ,
	\label{eq:classical-Bellman-operator-inf}
\end{align}
where $V =[V(s)]_{s\in \cS}$ with  $V(s) \coloneqq \max_a Q(s, a)$.
We propose to penalize the original Bellman operator w.r.t.~the empirical kernel $\widehat{P}$ as follows: for all $(s, a)\in\cS\times\cA$, set 
\begin{align}
	\Tpess (Q)(s, a) \coloneqq \max\Big\{r(s, a) + \gamma\widehat{P}_{s, a} V - b(s, a; V) , 0\Big\},
	\label{eq:empirical-Bellman-infinite}
\end{align}
%
where  $b(s, a; V)$ denotes the penalty term employed to enforce pessimism amid uncertainty \citep{jin2021pessimism,rashidinejad2021bridging}. 
As one can anticipate, the properties of the fixed point of $\Tpess(\cdot)$ relies heavily upon the choice of the penalty terms $\{b_h(s,a; V)\}$, often derived based on certain concentration bounds. Here, we focus on the following Bernstein-style penalty to exploit the importance of certain variance statistics:  
\begin{align}
	b(s, a; V) \defn \min \Bigg\{  \max \bigg\{ & \sqrt{\frac{\cb\log\frac{N}{(1-\gamma)\delta}}{N(s, a)}\mathsf{Var}_{\widehat{P}_{ s, a}} (V )} ,  \,\frac{2\cb \log\frac{N}{(1-\gamma)\delta}}{(1-\gamma)N(s, a)} \bigg\} ,\, \frac{1}{1-\gamma}  \Bigg\} +\frac{5}{N} 
	\label{def:bonus-Bernstein-infinite}
\end{align}
for every $(s,a)\in \cS\times \cA$, where $\cb> 0$ is some numerical constant (e.g., $\cb=144$), and $\delta \in (0,1)$ is some given quantity (in fact, $1-\delta$ is the target success probability).
Note that Bernstein-style UCBs add the term $b(s, a; V)$ as a bonus term to the estimated $Q$-function in order to promote exploration. 
Here, for any vector $V \in \mathbb{R}^S$, 
we recall that $\mathsf{Var}_{\widehat{P}_{s,a}} (V)$ is the variance of $V$ w.r.t.~the distribution $\widehat{P}_{s,a}$ (see \eqref{eq:defn-Var-P-V}).

\begin{algorithm}[t]
	\begin{algorithmic}[1]
	\STATE\textbf{input:} dataset $\mathcal{D}$; reward function $r$; target success probability $1-\delta$; max iteration number $\tau_{\max}$. \\
	\STATE\textbf{initialization:} $\widehat{Q}_{0}=0$, $\widehat{V}_0=0$. 
%
%
%

\FOR{$\tau =1,2,\cdots, \taumax $}
		\FOR{$s\in \cS, a\in \cA$}
		\STATE{ compute the penalty term $b\big(s,a; \widehat{V}_{\tau-1}\big)$ according to \eqref{def:bonus-Bernstein-infinite}. }
		\STATE{ set $\widehat{Q}_{\tau}(s, a) = \max\big\{r(s, a) + \gamma\widehat{P}_{s, a}\widehat{V}_{\tau-1} - b\big(s, a; \widehat{V}_{\tau-1}\big), 0\big\}$.}
	\ENDFOR
		\FOR{$s\in \cS$}
		\STATE{ set $\widehat{V}_{\tau}(s) = \max_a \widehat{Q}_{\tau}(s,a)$. \label{alg:infinite-q-update}}
	\ENDFOR
	\ENDFOR

	\STATE\textbf{output:} $\widehat{\pi}$ s.t.~$\widehat{\pi}(s) \in \arg\max_a \widehat{Q}_{\taumax}(s,a)$ for any $s\in \cS$. 
		\end{algorithmic}
	\caption{Offline value iteration with LCB (VI-LCB)}
 \label{alg:vi-lcb-infinite}
\end{algorithm}

\paragraph{The VI-LCB algorithm.} We are now positioned to introduce the VI-LCB algorithm, which can be regarded as classical value iteration applied in conjunction with pessimism.  Specifically, the algorithm applies the pessimistic operator $\Tpess$ (cf.~\eqref{eq:empirical-Bellman-infinite}) iteratively in order to find its fixed point:    
%
\begin{align}
	\widehat{Q}_{\tau}(s, a) & = \Tpess\big(\widehat{Q}_{\tau-1}\big)(s, a) \nonumber \\
	& = \max\Big\{r(s, a) + \gamma\widehat{P}_{s, a}\widehat{V}_{\tau-1} - b\big(s, a; \widehat{V}_{\tau-1} \big), 0\Big\},
	\qquad \tau = 1,2,\cdots
	\label{eq:VI-LCB-iterations-basic-inf}
\end{align}
We shall initialize it to $\widehat{Q}_0=0$, implement \eqref{eq:VI-LCB-iterations-basic-inf} for $\tau_{\max}$ iterations, 
and output $\widehat{Q}= \widehat{Q}_{\tau_{\max}}$ as the final Q-estimate. 
The final policy estimate $\widehat{\pi}$ is chosen on the basis of $\widehat{Q}$ as follows:  
\begin{align}
	\widehat{\pi}(s) \in \arg\max_a \widehat{Q}(s, a) \qquad \text{for all }s\in \cS,
\end{align}
with the whole algorithm summarized in Algorithm~\ref{alg:vi-lcb-infinite}.  


Let us pause to explain the rationale of the pessimism principle on a high level. If a pair $(s,a)$ has been insufficiently visited in $\mathcal{D}$ (i.e., $N(s,a)$ is small), 
then the resulting Q-estimate $\widehat{Q}_{\tau}(s, a)$ could suffer from high uncertainty and become unreliable, which might in turn mislead value estimation. 
By enforcing suitable penalization $b(s,a; \widehat{V}_{\tau-1})$ based on certain lower confidence bounds, 
we can suppress the negative influence of such poorly visited state-action pairs. 
Fortunately, suppressing these state-action pairs might not result in significant bias in value estimation when $\Cstar$ is small;
for instance,  when the behavior policy $\pib$ resembles $\pi^{\star}$,  the poorly visited state-action pairs correspond primarily to suboptimal actions (as they are not selected by $\pi^{\star}$), making it acceptable to neglect these pairs.

Thanks to the $\gamma$-contraction property of $\Tpess$, 
the iterates $\{\widehat{Q}_{\tau}\}_{\tau\geq 0}$ produced by Algorithm~\ref{alg:vi-lcb-infinite} converge linearly to the fixed point  $\widehat{Q}_{\mathsf{pe}}^{\star}$, 
as asserted below. 

\begin{lemma} \label{lem:monotone-contraction}
	Suppose $\widehat{Q}_0=0$. Then the iterates of Algorithm~\ref{alg:vi-lcb-infinite} obey
	\begin{align}
		\widehat{Q}_{\tau} \leq \widehat{Q}_{\mathsf{pe}}^{\star} 
		\qquad \text{and}\qquad
		\big\|\widehat{Q}_{\tau} - \widehat{Q}_{\mathsf{pe}}^{\star} \big\|_{\infty} \le  \frac{\gamma^{\tau}}{1-\gamma} 
		\qquad\quad
		\text{for all }\tau \geq 0,
	\end{align}
	where $\widehat{Q}_{\mathsf{pe}}^{\star}$ is the unique fixed point of $\Tpess$. 
	As a consequence, by choosing $\taumax \geq \frac{\log\frac{N}{1-\gamma}}{\log (1/\gamma)}$ one fulfills
	\begin{align}
		\big\|\widehat{Q}_{\taumax} - \widehat{Q}_{\mathsf{pe}}^{\star} \big\|_{\infty} \le  1/N. 
		\label{eq:taumax-Q-converge}
	\end{align}
\end{lemma}

When the Bernstein-style concentration bound \eqref{def:bonus-Bernstein-infinite} is adopted, 
the VI-LCB algorithm in Algorithm~\ref{alg:vi-lcb-infinite} yields $\varepsilon$-accuracy with a near-minimal number of samples, 
as stated below.  
\begin{theorem}[\cite{li2022settling}] \label{thm:infinite}
Suppose  $\gamma \in [\frac{1}{2},1)$, and consider any $0<\delta <1$ and $ \varepsilon \in \big(0, \frac{1}{1-\gamma} \big]$. 
Suppose that the total number of iterations exceeds $\taumax \geq \frac{1}{1-\gamma}\log\frac{N}{1-\gamma}$. 
With probability at least $1-2\delta,$  the policy $\widehat{\pi}$ returned by Algorithm~\ref{alg:vi-lcb-infinite} obeys
\begin{align}
	V^{\star}(\rho) - V^{\widehat{\pi}}(\rho) \leq \varepsilon,
\end{align}
provided that $\cb$ (cf.~the Bernstein-style penalty term in \eqref{def:bonus-Bernstein-infinite}) is some sufficiently large numerical constant and the total sample size exceeds 
\begin{align}
	N\geq\frac{c_1 S\Cstar \log\frac{NS }{(1-\gamma)\delta}}{(1-\gamma)^{3}\varepsilon^{2}} 
	\label{eq:N-range-epsilon2-inf-thm}
\end{align}
for some large enough numerical constant $c_1>0$, 
where $\Cstar$ is introduced in Definition~\ref{assumption:concentrate-infinite-simple}. 
\end{theorem}
%

In general, the total sample size characterized by Theorem~\ref{thm:infinite} could be far smaller than the ambient dimension (i.e., $S^2A$) of the transition kernel $P$, thus precluding one from estimating $P$ in a reliable fashion. 
As a crucial insight from Theorem~\ref{thm:infinite},  the model-based (or plug-in) approach enables reliable offline learning even when model estimation is completely off.  
Before further discussing implications of the above theorem, we develop a matching minimax lower bound that helps benchmark the efficacy of the developed algorithms.



%
%

%
\begin{theorem}[\cite{li2022settling}]\label{thm:infinite-lwoer-bound}
For any $(\gamma,S,\Cstar,\varepsilon)$ obeying $\gamma\in\big[\frac{2}{3},1\big),$ $S\geq 2$, $\Cstar\geq\frac{8\gamma}{S}$,
and $\varepsilon\leq\frac{1}{42(1-\gamma)}$, one can construct two
MDPs $\mathcal{M}_{0},\mathcal{M}_{1}$, an initial state distribution
$\rho$, and a batch dataset with $N$ independent samples and single-policy
clipped concentrability coefficient $\Cstar$ such that
\[
	\inf_{\widehat{\pi}}\max\left\{ \mathbb{P}_{0}\big( V^{\star}(\rho)-V^{\widehat{\pi}}(\rho)>\varepsilon\big), \,
	\mathbb{P}_{1}\big( V^{\star}(\rho) - V^{\widehat{\pi}}(\rho) >\varepsilon\big)\right\} \geq\frac{1}{8},
\]
provided that $$N\leq\frac{c_2S\Cstar}{(1-\gamma)^{3}\varepsilon^{2}}$$ 
for some numerical constant $c_{2}>0$. 
Here, the infimum is over all estimator $\widehat{\pi}$, and  
$\mathbb{P}_{0}$ (resp.~$\mathbb{P}_{1}$) denotes the probability
when the MDP is $\mathcal{M}_{0}$ (resp.~$\mathcal{M}_{1}$). 
\end{theorem}

Theorem~\ref{thm:infinite-lwoer-bound} suggests that no algorithm succeeds in finding the $\varepsilon$-optimal policy unless the sample size is above the order
$$ \Omega\left( \frac{S\Cstar}{(1-\gamma)^{3}\varepsilon^{2}} \right) .$$

\paragraph{Implications.} 
In what follows, we take a moment to interpret the above two theorems and single out several key implications about the proposed model-based algorithm.

\begin{itemize}
	\item 

{\em Optimal sample complexities.} 
In the presence of the Bernstein-style penalty, the total number of samples needed for VI-LCB to yield $\varepsilon$-accuracy is 
\begin{equation}
	\widetilde{O}\bigg( \frac{ S\Cstar  }{(1-\gamma)^3\varepsilon^2}  \bigg). 
	\label{eq:sample-size-Bernstein-infinite}
\end{equation}
This taken together with the minimax lower bound  in Theorem~\ref{thm:infinite-lwoer-bound} 
confirms its optimality  (up to some logarithmic factor). 


	\item
		{\em No burn-in cost.} The fact that the sample size bound \eqref{eq:N-range-epsilon2-inf-thm} holds for the full $\varepsilon$-range (i.e., any given $\varepsilon \in \big( 0, \frac{1}{1-\gamma} \big]$) means that there is no burn-in cost required to achieve sample optimality. 		

\end{itemize}

\section{Policy optimization}

Another popular approach for finding the optimal policy is to formulate the value maximization task as an optimization problem over the space of (parameterized) policies, and then attempt to solve it using optimization algorithms like policy gradient type methods. Such a policy-based approach is often referred to as policy optimization, which we briefly describe in this section.

\subsection{Policy gradient methods}

Given some prescribed initial distribution $\rho$ over $\cS$, policy optimization methods seek to maximize the value function
\begin{equation}\label{eq:policy_optimization}
\max_{\theta}\; V^{\pi}(\rho) \coloneqq V^{\pi_\theta}(\rho)
\end{equation}
over the policy $\pi: = \pi_\theta$  parameterized via some parameter $\theta$, with two examples given below. 
\begin{itemize}
    \item \textit{Direct parameterization:} the policy is directly parameterized by
$ \pi_\theta(a|s) = \theta(s,a)$,
    where  $\theta \in  \{\theta \in \mathbb{R}^{SA}:  \, \, \theta(s,a)\ge 0, \sum_{a\in\mathcal{A}}\theta(s,a)=1\}$.
    \item \textit{Tabular softmax parameterization:} For $\theta \in \mathbb{R}^{SA}$, the policy $\pi_\theta$ is generated through softmax transform
$\pi_\theta(a|s) = \frac{\exp(\theta(s,a))}{\sum_{a'\in\mathcal{A}}\exp(\theta(s,a'))}$,
    leading to an unconstrained optimization over $\theta$. This parameterization is also more popular with function approximation, by replacing $\theta(s,a)$ with a function approximator $f_\theta(s,a)$ implemented by neural networks. 
\end{itemize}

\paragraph{Policy gradients.} A natural approach is to apply first-order optimization methods (typically gradient-based methods) to solve the optimization problem \eqref{eq:policy_optimization}. Hence, the gradient $\nabla_\theta V^{\pi_\theta}(\rho)$, which is given by \citep{williams1992simple}
\begin{align*}
    \nabla_\theta V^{\pi_\theta}(\rho) & = \frac{1}{1-\gamma}\mathbb{E}_{s\sim d_\rho^{\pi_\theta}, a\sim \pi_\theta(\cdot|s)}\big[\nabla_\theta \log \pi_\theta (a|s) Q^{\pi_\theta}(s,a)\big]  \\
     & = \frac{1}{1-\gamma}\mathbb{E}_{s\sim d_\rho^{\pi_\theta}, a\sim \pi_\theta(\cdot|s)}\big[\nabla_\theta \log \pi_\theta (a|s) A^{\pi_\theta}(s,a)\big]  ,
\end{align*}
plays an instrumental role. 
The use of the advantage function $A^{\pi_\theta}(s,a) =Q^{\pi_\theta}(s,a) - V^{\pi_\theta}(s)$, rather than the Q-function $Q^{\pi_\theta}(s,a)$, often helps to reduce the variance of the policy gradients when sample-based estimates are employed, such as via REINFORCE \citep{williams1992simple}. 

For notational simplicity, we shall denote by $\theta^{(t)}$ and $\pi^{(t)}$ the parameter and the policy at the $t$-th iteration, and use $V^{(t)}$, $Q^{(t)}$, $A^{(t)}$, $d_\rho^{(t)}$ to denote $V^{\pi^{(t)}}$, $Q^{\pi^{(t)}}$, $A^{\pi^{(t)}}$, $d_\rho^{\pi^{(t)}}$, respectively. In addition, we assume the policy gradients and the value functions are exactly evaluated throughout this article, which enables us to focus on the optimization aspect of PG methods.  

\paragraph{Projected policy gradient method.}
Perhaps the most straightforward first-order policy optimization method is to adopt direct parameterization and perform projected gradient ascent updates:
\begin{equation}
    \theta^{(t+1)} = \mathcal{P}_{\Delta(\mathcal{A})^{S}}\big(\theta^{(t)} + \eta \nabla_\theta V^{(t)}(\rho) \big),
    \label{eq:direct_pg}
\end{equation}
where $\eta>0$ is the learning rate, $\mathcal{P}_{\Delta(\mathcal{A})^{S}}$ indicates the projection operator, and
\begin{equation*}
    \nabla_{\theta(s,a)} V^{(t)}(\rho) = \nabla_{\pi(s,a)} V^{(t)}(\rho) = \frac{1}{1-\gamma} d_\rho^{(t)}(s)Q^{(t)}(s,a).
\end{equation*}
As the value function $V^{\pi_\theta}(\rho)$ is $\frac{2\gamma A}{(1-\gamma)^3}$-smooth \citep{agarwal2019optimality}, setting the learning rate to $0 < \eta \le \frac{(1-\gamma)^3}{2\gamma A}$ ensures monotonicity of $V^{(t)}(\rho)$ in  $t$. In addition, it has been established in \citet{agarwal2019optimality} that the value function satisfies a gradient domination condition that associates the optimality gap $V^\star(\rho) - V^{\pi}(\rho)$ with a variational gradient term, thereby allowing the iterates to converge globally, as stated below.
\begin{theorem}[\mbox{\cite{agarwal2019optimality}}] With $0 < \eta \le \frac{(1-\gamma)^3}{2\gamma A}$, the iterates of the projected PG method \eqref{eq:direct_pg} satisfies
\begin{equation*}
    \min_{0\le t \le T} V^\star (\rho) - V^{(t)}(\rho) \le \frac{4\sqrt{S}}{1-\gamma}\bigg\|\frac{d_\rho^{\pi^{\star}}}{\rho}\bigg\|_\infty \sqrt{\frac{2(V^\star (\rho) - V^{(0)}(\rho))}{\eta T}}.
\end{equation*}
\label{thm:agarwal_pgd}
\end{theorem}
Theorem \ref{thm:agarwal_pgd} establishes an iteration complexity of $\mathcal{O}\Big(\frac{SA}{(1-\gamma)^6\varepsilon^2}\big\|\frac{d_\rho^{\pi^{\star}}}{\rho}\big\|_\infty^2\Big)$ for finding an $\varepsilon$-optimal policy, which is later improved to $\mathcal{O}\Big(\frac{SA}{(1-\gamma)^5\varepsilon}\big\|\frac{d_\rho^{\pi^{\star}}}{\rho}\big\|_\infty^2\Big)$ \citep{xiao2022convergence}. However, the projection operator introduces $\mathcal{O}(\log A)$ computational overhead every iteration and is less compatible with  function approximation. This motivates the studies of PG methods that are compatible with unconstrained optimization, e.g., by using softmax parameterization.


\paragraph{Softmax policy gradient method.} With softmax parameterization, the policy gradient method can be expressed as
\begin{equation}
    \theta^{(t+1)} = \theta^{(t)} + \eta\nabla_\theta V^{(t)}(\rho),
    \label{eq:softmax_pg}
\end{equation}
where
\begin{equation*}
    \nabla_{\theta(s,a)} V^{(t)}(\rho) = \frac{\eta}{1-\gamma}d_\rho^{(t)}(s)\pi^{(t)}(a|s)A^{(t)}(s,a).
\end{equation*}
Remarkably, \citet{agarwal2019optimality} established the asymptotic global convergence of the softmax PG method as follows.
\begin{theorem}[\mbox{\cite{agarwal2019optimality}}]
    With constant learning rate $0 < \eta \le (1-\gamma)^3/8$, the softmax PG method converges to the optimal policy, i.e., $V^{(t)}(s) \to V^\star(s)$ as $t\to \infty$ for all $s \in \cS$.
\end{theorem}


\citet{mei2020global} later demonstrated an iteration complexity of $\mathcal{O}\big(\frac{1}{c(\mathcal{M})^2\varepsilon}\big)$ for achieving an $\varepsilon$-optimal policy, where $c(\mathcal{M})$ is a trajectory-dependent quantity depending on salient problem parameters such as the number of states $S$ and the effective horizon $(1-\gamma)^{-1}$. Unfortunately, this quantity $c(\mathcal{M})$ can be extremely small and does not exclude the possibility of incurring excessively large iteration complexity, as demonstrated by the following hardness result \citep{li2021softmax}.
\begin{theorem}[\cite{li2021softmax}]
    There exist universal constants $c_1, c_2, c_3 > 0$ such that for any $\gamma \in (0.96, 1)$ and $S \ge c_3 (1-\gamma)^{-6}$, one can find a $\gamma$-discounted MDP such that the softmax PG method takes at least
    \begin{equation*}
        \frac{c_1}{\eta} S^{2^{\frac{c_2}{1-\gamma}}}
    \end{equation*}
    iterations to reach $\|V^\star - V^{(t)}\|_\infty \le 0.15$.
\end{theorem}
Therefore, while guaranteed to converge globally, softmax PG methods can take (super-)exponential time to even reduce the optimality gap to within a constant level. 
Intuitively speaking, softmax PG methods fail to achieve a reasonable convergence rate when the probability $\pi^{(t)}(a^\star(s)\,|\,s)$ assigned to the optimal action $a^\star(s)$ is too close to zero. 

\paragraph{Natural policy gradient method.} Both projected PG and softmax PG methods fall short of attaining an iteration complexity that is independent of salient problem parameters, especially with respect to the size of the state space $S$. This ambitious goal can be achieved, somewhat surprisingly, by adopting the Fisher information matrix as a preconditioner, which leads to the natural policy gradient (NPG) method \citep{kakade2002natural}:
\begin{equation}    \label{eq:softmax_npg}
        \theta^{(t+1)} = \theta^{(t)} + \eta (\mathcal{F}_\rho^{\theta^{(t)}})^\dagger \nabla_\theta V^{(t)}(\rho),
\end{equation}
where
$   \mathcal{F}_\rho^\theta := \mathbb{E}_{s\sim d_\rho^{\pi_\theta}, a\sim \pi_\theta(\cdot|s)}\Big[\big(\nabla_\theta\log \pi_\theta(a|s)\big)\big(\nabla_\theta\log \pi_\theta(a|s)\big)^\top\Big]$
is the Fisher information matrix,  and $^\dagger$ denotes the Moore-Penrose pseudoinverse. With softmax parameterization, the NPG updates take the form 
\begin{equation*}
    \theta^{(t+1)} = \theta^{(t)} + \frac{\eta}{1-\gamma}A^{(t)},
\end{equation*}
or equivalently,
\begin{equation*}
    \pi^{(t+1)}(a|s) ~\propto~ \pi^{(t)}(a|s) \exp\Big(\frac{\eta Q^{(t)}(s,a)}{1-\gamma}\Big).
\end{equation*}
Note that the (softmax) NPG update rule coincides with the multiplicative weights update (MWU) method \citep{cesa2006prediction}, and that the update rule does not depend on the initial state distribution $\rho$. \citet{shani2019adaptive} first established a global convergence rate of $\mathcal{O}\big(\frac{1}{(1-\gamma)^2\sqrt{T}}\big)$ using diminishing learning rates $\eta_t = \mathcal{O}\big(\frac{1-\gamma}{\sqrt{t}}\big)$, which was subsequently improved by \citet{agarwal2019optimality} using a constant learning rate $\eta$, stated below.
\begin{theorem}[\mbox{\cite{agarwal2019optimality}}]     \label{thm:npg}
    With uniform initialization $\theta^{(0)} = 0$ and constant learning rate $\eta > 0$, the iterates of NPG satisfy
    \begin{equation*}
        V^\star(\rho) - V^{(T)}(\rho) \le \frac{1}{T}\Bigg(\frac{\log A }{\eta} + \frac{1}{(1-\gamma)^2}\Bigg).
    \end{equation*}
\end{theorem}
Encouragingly, as long as $\eta\geq \frac{(1-\gamma)^2}{\log A }$, the iteration complexity of NPG methods becomes $\mathcal{O}\big(\frac{1}{(1-\gamma)^2 T}\big)$, which is independent of the size of the state-action space. 
On the complementary side, the iteration complexity of NPG is lower bounded by $\frac{\Delta}{(1-\gamma) A }\exp(-\eta\Delta T)$ --- established in \citet{khodadadian2021linear} --- where the optimal advantage function gap $\Delta  = \min_s\min_{a\neq a^{\star}(s)} |A^{\star}(s,a)| \ge 0$ is determined by the MDP instance. As the lower bound attains its maximum $\frac{1}{(1-\gamma)e\eta T}$ when $\Delta = \frac{1}{\eta T}$, it is immediate that the sublinear rate in Theorem \ref{thm:npg} cannot be improved in $T$. Nonetheless, two strategies to achieve even faster linear convergence with NPG updates include (i) adopting increasing/adaptive learning rates \citep{khodadadian2021linear,bhandari2020note,lan2021policy,xiao2022convergence}, or (ii) introducing entropy regularization \citep{cen2020fast,lan2021policy,zhan2021policy}, as discussed next.

\subsection{Entropy regularization} 
Entropy regularization is a popular technique in practice to promote exploration \citep{williams1991function}. Specifically, instead of optimizing the original value function,  one seeks to optimize the entropy-regularized value function defined as
\begin{equation*}
        V_\tau^{\pi}(s) = V^{\pi}(s) + \frac{\tau}{1-\gamma} \mathbb{E}_{s'\sim d_s^{\pi}}\big[ \mathcal{H} (\pi(\cdot|s'))\big],
\end{equation*}
where $\mathcal{H}(\pi(\cdot|s)) = - \sum_{a\in \mathcal{A}} \pi(a|s) \log \pi(a|s)$ is the entropy of policy $ \pi(\cdot |s) $, and $\tau > 0$ serves as the regularization parameter known as the temperature. The entropy-regularized $Q$-function is defined as
$$        Q_\tau^{\pi}(s,a) = r(s,a) + \gamma  \mathbb{E}_{s'\sim P(\cdot|s,a)} \big[ V_\tau^{\pi}(s') \big]. $$
The resulting optimal value function, $Q$-function and optimal policy are denoted by $V_\tau^\star$, $Q_\tau^\star$, and $\pi_\tau^{\star}$, respectively.
From an optimization perspective, the entropy term adds curvature to the value function and ensures that the optimal policy $\pi_\tau^\star$ is unique. Interestingly, in contrast to the greedy optimal policy for the unregularized problem in \eqref{eq:policy_optimization}, the optimal policy of the entropy-regularized problem reflects ``bounded rationality'' in decision making \citep{mckelvey1995quantal}, namely
$$ \pi_\tau^\star (\cdot| s)~\propto~ \exp\left( Q_\tau^{\star}(s,\cdot) /\tau \right) .$$
It should be noted, however, adding the entropy regularization generally does not make $V_\tau^\pi(\rho)$ concave unless $\tau$ is unreasonably large.  As the entropy function is bounded by $  \log|\mathcal{A}|$, the optimal entropy-regularized policy is also guaranteed to be approximately optimal for the unregularized RL problem in the following sense:
\begin{equation*}
    V^{\pi_\tau^\star}(\rho) \ge V^\star(\rho) - \frac{\tau\log A}{1-\gamma}.
\end{equation*}
  
Motivated by its benign convergence, we consider NPG for the entropy-regularized problem:
\begin{equation*}
    \theta^{(t+1)} = \theta^{(t)} + \eta (\mathcal{F}_\rho^{\theta^{(t)}})^\dagger \nabla_\theta V_\tau^{(t)}(\rho),
\end{equation*}
which can be equivalently written as
\begin{equation}     \label{eq:ent_NPG}
    \pi^{(t+1)}(a|s) ~\propto~ \pi^{(t)}(a|s)^{1-\frac{\eta\tau}{1-\gamma}} \exp\Big(\frac{\eta Q_\tau^{(t)}(s,a)}{1-\gamma}\Big)
\end{equation}
 under the softmax parameterization. The following theorem shows that with appropriate choices of constant learning rate $\eta$, entropy-regularized NPG converges to the unique optimal policy $\pi_\tau^\star$ at a linear rate.
\begin{theorem}[\mbox{\cite{cen2020fast,lan2021policy}}]    
    For constant learning rate $0 < \eta \le (1-\gamma)/\tau$ and uniform initialization, the entropy-regularized NPG updates \eqref{eq:ent_NPG} satisfy
    \begin{equation*}
        \|V_\tau^\star - V_\tau^{(T)}\|_\infty \le \frac{15(1 + \tau \log A)}{1-\gamma} (1-\eta\tau)^{T-1}
    \end{equation*}
and 
    \begin{align*}
        V_\tau^\star(\rho) - V_\tau^{(T)}(\rho) & \le \bigg\|\frac{\rho}{\nu_\tau^\star}\bigg\|_\infty\bigg(\frac{1+\tau\log A}{1-\gamma}+\frac{(1-\gamma)\log A}{\eta}\bigg)     \max\Big\{\gamma, 1-\frac{\eta\tau}{1-\gamma}\Big\}^T.
    \end{align*}
 Here, $\nu_\tau^\star$ is the stationary state distribution of policy $\pi_\tau^\star$.
\end{theorem}
The first and the second bounds are due to \citet{cen2020fast} and \citet{lan2021policy}\footnote{We discard some of the simplification steps therein and state the convergence result for a wider range of learning rate $\eta$.} respectively, where they lead to slightly different iteration complexities. Taken collectively, entropy-regularized NPG takes no more than
\begin{equation*}
    \widetilde{\mathcal{O}}\Big(\min\Big\{\frac{1}{\eta\tau}\log\frac{1}{\varepsilon},\, \max\Big\{\frac{1}{1-\gamma}, \frac{1-\gamma}{\eta\tau}\Big\}\log\frac{\norm{{\rho}/{\nu_\tau^\star}}_\infty}{\varepsilon}\Big\}\Big) 
\end{equation*}
iterations to find a policy satisfying $V_\tau^\star(\rho) - V_\tau^\pi(\rho) \le \varepsilon$.
We make note that the difference stems from different analysis approaches: \citet{cen2020fast} built their analysis upon the contraction property of the soft Bellman operator (the entropy-regularized counterpart of the original Bellman operator), while \citet{lan2021policy} made use of the connection between regularized NPG and regularized mirror descent. This can be observed from the following equivalence: the update rule \eqref{eq:ent_NPG} can be equivalently expressed as
\ifdefined\arxiv
\begin{align*}
     \pi^{(t+1)}(\cdot|s)  &= \arg\min_{p\in\Delta(\cA)} \innprod{p, -Q_\tau^{(t)}(s,\cdot)} - \tau \mathcal{H}(p)   + \frac{1}{\eta_{\texttt{MD}}}\KL{p}{\pi^{(t)}(\cdot|s)},
\end{align*}
\else
\begin{align*}
     \pi^{(t+1)}(\cdot|s)  &= \arg\min_{p\in\Delta(\cA)} \innprod{p, -Q_\tau^{(t)}(s,\cdot)} - \tau \mathcal{H}(p)  + \frac{1}{\eta_{\texttt{MD}}}\KL{p}{\pi^{(t)}(\cdot|s)},
\end{align*}
\fi
with $\eta_{\texttt{MD}} = \frac{\eta}{1-\gamma-\eta\tau}$. The analysis of regularized RL can be further generalized to adopt non-strongly convex regularizers \citep{lan2021policy}, non-smooth regularizers \citep{zhan2021policy}, Markov games \citep{cen2024fast,cen2023faster}, potential games \citep{cen2022independent}, and so on.

\section{Distributionally robust RL}

While standard RL has been heavily investigated, its use can be significantly hampered in practice due to the sim-to-real gap or uncertainty \citep{bertsimas2019adaptive}; 
for instance, a policy learned in an ideal, nominal environment might fail catastrophically when the deployed environment is subject to small changes in task objectives or adversarial perturbations \citep{zhang2020robust,ding2023seeing}. Consequently, in addition to maximizing the long-term cumulative reward, robustness emerges as another critical goal for  RL, especially in high-stakes applications such as robotics, autonomous driving, clinical trials, financial investments, and so on. Towards achieving this, distributionally robust RL \citep{iyengar2005robust,nilim2005robust,xu2012distributionally,shi2024sample,shi2025breaking}, which leverages insights from distributionally robust optimization and supervised learning \citep{gao2023finite,bertsimas2018data,duchi2018learning,blanchet2019quantifying}, becomes a natural yet versatile framework;   
 the aim is to learn a policy that performs well even when the deployed environment deviates from the nominal one in the face of environment uncertainty.


\subsection{Distributionally robust MDPs}

We now introduce the distributionally robust MDP (RMDP) tailored to the infinite-horizon discounted setting, 
denoted by 
$$\cM_{\mathsf{rob}} = \left\{\cS,\cA, \gamma, \cU_\rho^{\ror}(P^\no), r \right\}, $$ 
where $\cS, \cA, \gamma, r$ are identical to those in the standard MDP introduced in Section~\ref{chap:mdp}. A key distinction from the standard MDP is that: rather than assuming a fixed transition kernel $P$, it allows the transition kernel to be chosen arbitrarily from a prescribed uncertainty set  $\cU_\rho^{\ror}(P^\no)$ centered around a {\em nominal} kernel $P^\no: \cS\times\cA \rightarrow \Delta(\cS)$, where the uncertainty set is specified using some distance metric $\rho$ of radius $\ror>0$. 
In particular, given the nominal transition kernel $P^\no$ and some uncertainty level $\ror$, the uncertainty set---with the divergence metric $\rho: \Delta(\cS) \times \Delta(\cS) \rightarrow \mathbb{R}^+$---is specified as
\begin{align}\label{eq:general-infinite-P}
	\cU_\rho^{\ror}(P^\no) \defn \otimes \; \cU_\rho^{\ror}(P^{\no}_{s,a}) \qquad 
\mbox{with} \qquad  
	\cU_\rho^{\ror}(P^\no_{s,a}) \defn \left\{ P_{s,a} \in \Delta (\cS): \rho \left(P_{s,a}, P^0_{s,a}\right) \leq \ror \right\},
\end{align}
where we denote a vector of the transition kernel $P$ or $P^{\no}$ at state-action pair $(s,a)$ respectively as
\begin{align}\label{eq:defn-P-sa}
	P_{s,a} \defn P(\cdot \mymid s,a) \in \mathbb{R}^{1\times S}, \qquad P_{s,a}^\no \defn P^\no(\cdot \mymid s,a) \in \mathbb{R}^{1\times S}.
\end{align}
In other words, the uncertainty is imposed in a decoupled manner for each state-action pair to ensure  tractability, obeying the so-called $(s,a)$-rectangularity \citep{zhou2021finite,wiesemann2013robust,wang2023foundation}. Compared with standard MDPs, the class of RMDPs encapsulates richer models, 
given that one is allowed to prescribe the shape and size of the uncertainty set. For ease of exposition, we consider a popular choice of the uncertainty set measured in terms of the total variation distance, given by 
\begin{align}\label{eq:tv-distance}
\rho_{\TV}\left(P_{s,a}, P^0_{s,a}\right) & \defn  \frac{1}{2} \sum_{s'\in \cS} P^{\no}(s' \mymid s,a)\left| 1 - \frac{P(s' \mymid s,a)}{P^{\no}(s' \mymid s,a)}\right|.   
\end{align}
Note that $\rho_{\TV}\left(P_{s,a}, P^0_{s,a}\right)\in [0,1]$. Other choices of divergence metrics are also investigated, such as the $\chi^2$ distance \citep{yang2021towards,shi2024curious}, the $L_p$ distance \citep{clavier2024near}, and the KL divergence \citep{panaganti2022sample,shi2022distributionally}.

In RMDPs, we are interested in the worst-case performance of a policy $\pi$ over all the possible transition kernels in the uncertainty set. This is measured by the {\em robust value function} $V^{\pi, \ror} $ and the {\em robust Q-function} $Q^{\pi,\ror}$ in $\cM_\rob$, defined respectively as
\begin{align} \label{eq:robust-value-def}
\forall (s,a)\in \cS \times \cA: \quad  V^{\pi,\ror}(s) &\defn \inf_{P\in \unb_\rho^{\ror}(P^{\no})} V^{\pi,P} (s),  \quad  
	  Q^{\pi,\ror}(s,a)  \defn \inf_{P\in \unb_\rho^{\ror}(P^{\no})} Q^{\pi,P}(s,a).
\end{align}

\paragraph{Optimal robust policy and robust Bellman operator.}
As a generalization of standard MDPs, it follows that there exists at least one deterministic policy that maximizes the robust value function (resp.~robust Q-function) simultaneously for all states (resp.~state-action pairs) \citep{iyengar2005robust,nilim2005robust}. We denote the {\em optimal robust value function} (resp.~{\em optimal robust Q-function}) as $V^{\star,\ror}$ (resp.~$Q^{\star,\ror}$), and the optimal robust policy as $\pi^\star$, which satisfy
\begin{subequations} \label{eq:mdp-value-Q}
\begin{align}
	\forall s \in \cS: \quad &V^{\star,\ror}(s) \defn V^{\pi^\star,\ror}(s) = \max_\pi V^{\pi,\ror}(s), \\
	\forall (s,a) \in \cS \times \cA: \quad &Q^{\star,\ror}(s,a) \defn Q^{\pi^\star,\ror}(s,a) = \max_\pi Q^{\pi,\ror}(s,a).
\end{align}
\end{subequations}
A key machinery in RMDPs is a generalization of Bellman's optimality principle, 
encapsulated in the following {\em robust Bellman optimality equation}:
%
\begin{align}
	\forall (s,a)\in \cS\times \cA: \quad &Q^{\star,\ror}(s,a) = r(s,a) + \gamma\inf_{\cP\in \unb^{\ror}_\rho(P^{\no}_{s,a})} \cP V^{\star,\ror}. \label{eq:bellman-equ-star-infinite}
\end{align}
%
The robust Bellman operator \citep{iyengar2005robust,nilim2005robust}---denoted by $\cT^\ror(\cdot): \mathbb{R}^{SA} \mapsto \mathbb{R}^{SA}$---is defined as follows: 
\begin{align}\label{eq:robust_bellman}
	\forall (s,a)\in \cS\times \cA :\quad \cT^\ror(Q)(s,a) \defn r(s,a) + \gamma \inf_{ \cP \in \unb^{\ror}_\rho(P^{\no}_{s,a})} \cP V,  \;\; \mbox{with}\; V(s) \defn \max_a Q(s,a).
	\end{align}
Given that $Q^{\star,\ror}$ is the unique fixed point of $\cT^\ror$,
one can recover the optimal robust value function and Q-function using a procedure termed {\em distributionally robust value iteration} (DRVI). 
Generalizing the standard value iteration, DRVI starts from some given initialization and recursively applies the robust Bellman operator until convergence. 
From an initialization $Q_0 = 0$, the update rule at the $t$-th ($t\geq 1$) iteration can be formulated as:
\begin{align}
	\forall (s,a)\in \cS\times \cA: \quad Q_t(s,a) &= \cT^\ror \big( {Q}_{t-1} \big)(s,a)  . \label{eq:vi-iteration}
\end{align}
However, directly solving \eqref{eq:vi-iteration} is computationally expensive
since it involves optimization over an $S$-dimensional probability simplex at each iteration, especially when the dimension of the state space $\cS$ is large. Fortunately, in view of strong duality \citep{iyengar2005robust}, \eqref{eq:vi-iteration} can be equivalently solved using its dual problem, which concerns optimizing a {\em scalar} dual variable and thus can be solved efficiently. In what follows, we shall illustrate this for the choice of the divergence $\rho$ of interest (cf.~\eqref{eq:tv-distance}). 
We have the following lemma enabling an equivalent dual update rule for \eqref{eq:vi-iteration} in DRVI.
 \begin{lemma}[Strong duality~\cite{iyengar2005robust}]\label{lemma:dual-form}
Consider any probability vector  $P\in\Delta(\cS)$, any fixed uncertainty level $\ror$ and the uncertainty set $\unb^{\ror}(P)$ specified with \eqref{eq:tv-distance}. For any vector $V\in \mathbb{R}^S$ obeying $  V \geq  {0}$, one has
\begin{align}
	\inf_{ \cP \in \unb^{\ror}(P)} \cP V  &= 
	 \max_{\alpha\in [ V_{\min},  V_{\max}]} \left\{P \left[V\right]_{\alpha} - \ror \left(\alpha - \min_{s'}\left[V\right]_{\alpha}(s') \right)\right\} ,  \label{eq:vi-l1norm} 
\end{align}
where $V_{\min} = \min_s V(s)$, $V_{\max} = \max_s V(s)$ and $[V]_\alpha = \min \{ V(s), \alpha \}$ for some $\alpha\geq 0$.
\end{lemma}
This procedure, again, converges rapidly due to the $\gamma$-contraction property of $\cT^\ror$  w.r.t.~the $\ell_\infty$ norm \citep{iyengar2005robust,nilim2005robust}.

\subsection{Model-based robust RL}

We assume access to a generative model or a simulator \citep{kearns1999finite}, which allows us to collect $N$ independent samples for each state-action pair 
generated based on the {\em nominal} kernel $P^{\no}$:
\begin{align}
	\forall (s,a)\in \cS\times\cA, \qquad s^{(i)}(s,a) \overset{\text{i.i.d.}}{\sim} P^\no(\cdot \mymid s,a), \qquad i = 1, 2,\cdots, N.
\end{align}

\paragraph{Goal.}
Given the collected samples, the task is to learn the robust optimal policy for the RMDP---w.r.t.~some prescribed uncertainty set $\cU^\ror(P^\no)$ around the nominal kernel---using as few samples as possible. Specifically, given some target accuracy level $\varepsilon>0$, the goal is to seek an $\varepsilon$-optimal robust policy $\widehat{\pi}$ obeying
\begin{align}
	\forall s\in \cS:\quad V^{\star, \sigma}(s) - V^{\widehat{\pi}, \sigma}(s) \leq \varepsilon.
\end{align}

\paragraph{Model-based algorithm.}
We consider a model-based approach tailored to RMDPs, which first constructs an empirical nominal transition kernel based on the collected samples, and then applies DRVI (or any other planning algorithm) to compute an optimal robust policy. The empirical nominal transition kernel $\widehat{P}^\no \in \mathbb{R}^{SA\times S}$ can be constructed on the basis of the empirical frequency of state transitions, i.e.,
\begin{align}
	\forall (s,a,s')\in \cS\times \cA \times \cS:\quad \widehat{P}^0(s'\mymid s,a) \defn  \frac{1}{N} \sum\limits_{i=1}^N \mathds{1} \big\{   s' = s^{(i)}(s,a) \big\},
	\label{eq:empirical-P-infinite}
\end{align}
which leads to an empirical RMDP $\widehat{\cM}_{\mathsf{rob}} = \{\cS,\cA, \gamma, \cU^{\ror}_\rho(\widehat{P}^\no), r\}$. 
   Running DRVI \eqref{eq:vi-iteration} on the empirical RMDP for $T$ iterations, we output the greedy policy of the final Q-estimate $\widehat{Q}_T$ as the final policy $\widehat{\pi}$, namely,
\begin{align}\label{alg:cvi-dro-infinite}
	 \forall s \in \cS: \quad \widehat{\pi}(s) = \arg\max_a \widehat{Q}_T(s,a).
\end{align}

\paragraph{Sample complexity of learning RMDPs.} The following theorem develops an upper bound on the sample complexity of DRVI in order to return an $\varepsilon$-optimal robust policy. 

\begin{theorem}[Upper bound under TV distance, \cite{shi2024curious}]\label{thm:l1-upper-bound} 
Consider any discount factor $\gamma \in \left[\frac{1}{4},1 \right)$, 
 uncertainty level $\ror\in (0,1)$, and $\delta \in (0,1)$. 
	Let $\widehat{\pi}$ be the output policy of \eqref{alg:cvi-dro-infinite} after $T = C_1 \log \big( \frac{N}{1-\gamma}\big)$ iterations. 
	Then with probability at least $1-\delta$, one has
\begin{align}
	\forall s\in\cS: \quad V^{\star, \ror}(s) - V^{\widehat{\pi}, \ror}(s) \leq \varepsilon
\end{align}
for any $\varepsilon \in \left(0, \sqrt{1/\max\{1-\gamma, \ror\}} \right]$,
as long as the total number of samples obeys
\begin{align}
	NSA \geq  \frac{C_2 SA}{ (1-\gamma)^2 \max\{1-\gamma, \ror\} \varepsilon^2}\log\left(\frac{SAN}{(1-\gamma)\delta}\right).
\end{align}
Here, $C_1, C_2>0$ are some large enough universal constants.  
\end{theorem}

%

Before discussing the implications of Theorem~\ref{thm:l1-upper-bound}, we note that a matching minimax lower bound is also established in \citet{shi2024curious}, confirming the tightness and optimality of the upper bound, which in turn pins down the sample complexity requirement for learning RMDPs with TV distance. Theorem~\ref{thm:l1-upper-bound} shows that the total number of samples required for DRVI (or any oracle planning algorithm) to yield $\varepsilon$-accuracy is
\begin{align}\label{eq:tv-final-samples}
\widetilde{O} \left(\frac{SA}{ (1-\gamma)^2 \max\{1-\gamma, \ror\} \varepsilon^2} \right).
\end{align}  
Recall that the sample complexity requirement for learning standard MDPs with a generative model is 
$ \widetilde{O} \left(\frac{SA}{ (1-\gamma)^3 \varepsilon^2} \right)$
in order to yield $\varepsilon$ accuracy \citep{agarwal2020model,li2020breaking}. 
Comparing this with the sample complexity requirement in \eqref{eq:tv-final-samples} for RMDPs under the TV distance, we confirm that the latter is at least as easy as---if not easier than---standard MDPs. In particular, when $\sigma \lesssim 1-\gamma$ is small, the sample complexity of RMDPs is the same as that of standard MDPs, which is as anticipated since the RMDP reduces to the standard MDP when $\sigma=0$. On the other hand, when $1-\gamma \lesssim \sigma < 1 $, the sample complexity of RMDPs simplifies to
$\widetilde{O} \left(\frac{SA}{ (1-\gamma)^2   \ror  \varepsilon^2} \right)$,
which is smaller than that of standard MDPs by a factor of $\sigma/(1-\gamma)$.

%
  
\paragraph{Discussion: other uncertainty sets.}  The sample complexity of RMDPs under other types of uncertainty sets is far much nuanced, and is not always smaller than that required by standard MDPs. For example, for the case of $\chi^2$ divergence, \citet{shi2024curious} suggested that RMDPs can be significantly harder than standard MDPs in certain ranges of the uncertainty level. It remains an interesting open question to establish tight sample complexities of RMDPs over broad families of uncertainty sets.

\section{Reinforcement learning with human feedback}

{\em Reinforcement learning from human feedback} (RLHF) \citep{ouyang2022training}, 
a paradigm to fine-tune large language models (LLMs), has been shown to be instrumental and significantly improve the helpfulness, truthfulness and controllability of LLMs. Roughly speaking, there are two critical components of RLHF: (1) {\em reward modeling}, which maps human preference rankings into a quantitative reward function that can guide policy improvement; and (2) {\em RL fine-tuning}, which seeks to adjust LLM output to align with human preferences by leveraging the learned reward function, i.e., increasing the probability of preferred answers and decreasing the probability of unfavored answers.  

Evidently, the curation of preference data is instrumental in the performance of RLHF, which is commonly modeled as pairwise comparisons from a Bradley-Terry ranking model \citep{bradley1952rank}.
In particular, given a query $x$, human annotators choose a preferred answer from two candidate answers $y_1$ and $y_2$ generated by an LLM. Despite the simple form, collecting large-scale and high-quality preference data can be expensive and time-consuming.
Depending on the availability of preference data, two paradigms of RLHF are considered: (1) \emph{offline} RLHF, where only a pre-collected preference dataset is available, possibly generated from a pre-trained LLM after supervised fine-tuning (SFT); and (2) \emph{online} RLHF, where additional preference data can be collected adaptively to improve alignment. 
How to best execute the principles of optimism and pessimism in the online and offline settings respectively---in a manner amenable with LLM-parameterized policies--- remains an area of active research.

\subsection{Background}

In RLHF, a language model is described by a policy $\pi$, which generates an answer $y \in \mathcal{Y}$ given prompt $x\in \mathcal{X}$ according to the conditional probability distribution $\pi(\cdot | x)$.\footnote{For simplicity of exposition, we assume a contextual bandit formulation, which can be regarded as an MDP with $H=1$.} The standard RLHF process consists of four stages: supervised fine-tuning (SFT), preference data generation, reward modeling, and RL fine-tuning. In the SFT stage, a language model $\pi_{\text{sft}}$ is obtained by fine-tuning a pre-trained LLM with supervised learning. The remaining stages continue training by leveraging the preference data, which we elaborate below.
 
\paragraph{Reward modeling from preference data.} An oracle (e.g., a human labeler or a scoring model) evaluates the quality of two answers $y_1$ and $y_2$ given prompt $x$ and reveals its preference. A widely used approach for modeling the probability of pairwise preferences is the Bradley-Terry model \citep{bradley1952rank}:
\begin{align} \label{eq:BT}
    \mathbb{P}(y_1\succ y_2|x) &= \frac{\exp(r^\star(x, y_1))}{\exp(r^\star(x, y_1))+\exp(r^\star(x, y_2))}   = \sigma(r^\star(x, y_1) - r^\star(x, y_2)),
\end{align}
where $y_1\succ y_2$ indicates that $y_1$ is preferred over $y_2$, $r^\star: \mathcal{X}\times\mathcal{Y}\to \mathbb{R}$ is the ground truth reward function, and $\sigma : \mathbb{R} \to (0, 1)$ is the logistic function. A preference data sample is denoted by a tuple $(x, y_+, y_-)$, where $y_+$ (resp.  $y_{-}$) is the preferred (resp. unpreferred) answer in the comparison.

Given a preference dataset $\mathcal{D} = \{(x^i, y_+^i, y_-^i)\}$ composed of independent samples, the reward function $r$ can be estimated by maximum likelihood estimation (MLE):
\begin{equation}
    r_{\mathsf{MLE}} = \arg\min_{r} \; \ell(r, \mathcal{D}),
    \label{eq:MLE}
\end{equation}
where $\ell(r, \mathcal{D})$ is the negative log-likelihood of $\mathcal{D}$, given as
\begin{equation}
     \ell(r, \mathcal{D}) \coloneqq -\sum_{(x^i, y_+^i, y_-^i)\in\mathcal{D}}{\log \sigma(r(x^i,y_+^i) - r(x^i,y_-^i))}.
\end{equation}

\paragraph{RL fine-tuning.} Given a reward model $r$, we seek to fine-tune the policy $\pi$ to achieve an ideal balance between the expected reward and its distance from an initial policy $\pi_{\text{ref}}$, which is typically the same as $\pi_{\text{sft}}$. This is  achieved by maximizing the KL-regularized value function $J(r, \pi)$, defined as
\begin{equation} \label{eq:KL_reward}
    J(r, \pi) = \exlim{x \sim \rho, y \sim \pi(\cdot|x)}{r(x, y) } - \beta\exlim{x \sim \rho}{ \KL{\pi(\cdot|x)}{\pi_{\text{ref}}(\cdot|x)}},
\end{equation}
where $\KL{\pi_1}{\pi_2} $ is the KL divergence from $\pi_1$ to $\pi_2$, and $\beta>0$ is a regularization parameter. Consequently, the RL fine-tuned policy $\pi_r$ with respect to the reward $r$ satisfies  
\begin{equation}
    \pi_r \coloneqq \arg\max_{\pi} J(r, \pi),
\end{equation}
which admits a closed-form solution \citep{rafailov2023direct}, i.e.,
\begin{equation}
\forall (x\times y) \in \mathcal{X}\times\mathcal{Y}: \qquad    \pi_r(y|x) = \frac{\pi_{\text{ref}}(y|x)\exp(r(x,y)/\beta)}{Z(r, x)}.
    \label{eq:RLHF-policy-closed-form}
\end{equation}
Here, $ Z(r, x)$ is a normalization factor given by
$  Z(r, x) = \sum_{y' \in \mathcal{Y}}\pi_{\text{ref}}(y'|x)\exp(r(x,y')/\beta)$.

\paragraph{Direct preference optimization.} 
The closed-form solution \eqref{eq:RLHF-policy-closed-form} allows us to write the reward function $r$ in turn as
\begin{equation}\label{eq:reward-policy-eq}
    r(x, y) = \beta (\log \pi_r(y|x) - \log \pi_{\text{ref}}(y|x) + \log Z(r, x)).
\end{equation}
Plugging the above equation into the reward MLE~\eqref{eq:MLE}, we obtain the seminal formulation of direct preference optimization (DPO) over the policy space \citep{rafailov2023direct}, 
\begin{align} \label{eq:DPO}
  \pi_{\mathsf{DPO}}  &= \arg\min_{\pi} \; -\sum_{(x^i, y_+^i, y_-^i)\in\mathcal{D}}{\log \sigma\left( \beta \left(\log \frac{\pi(y_+^i|x)}{\pi_{\text{ref}}(y_+^i|x)}  - \log \frac{\pi(y_-^i|x)}{\pi_{\text{ref}}(y_-^i|x)}  \right) \right)} ,
\end{align}
which avoids explicitly learning the reward model and is friendly to policy optimization.

\subsection{Value-incentivized preference optimization}

A major caveat of the standard RLHF framework concerns the lack of accounting for reward uncertainty, which is known to be indispensable in the success of standard RL paradigms in both online and offline settings. This motivates the design of a principled mechanism that be easily integrated into the RLHF pipeline, while bypassing the difficulties of explicit uncertainty estimation in LLMs.

In view of the sub-optimality of naive MLE for reward estimation \citep{cesa2017boltzmann,rashidinejad2021bridging}, and motivated by the effectiveness of reward-biased MLE in online RL \citep{kumar1982new,liu2019exploration}, we propose to regularize the reward estimate via
\begin{equation} 
J^\star(r) = \max_{\pi} \, J(r,\pi),
\end{equation}
which measures the resulting value function for the given reward if one acts according to its optimal policy. However, in RLHF,  by the definition \eqref{eq:BT}, the reward function $r^\star$ is only identifiable up to a prompt-dependent global shift. Specifically, letting $r_1(x, y) = r_2(x, y) + c(x)$ be two reward functions that only differ by a prompt-dependent shift $c(x)$, we have $r_1(x, y_1) - r_1(x, y_2) = r_2(x, y_1) - r_2(x, y_2)$, which leads to $J^{\star}(r_1) = J^{\star}(r_2) +\mathbb{E}_{x\sim \rho}[c(x)]$. To resolve this challenge, we introduce the following equivalent class of reward functions for the Bradley-Terry model to eliminate the shift ambiguity, which also has the calibration effect of centering the reward function while offering a regularization mechanism to incorporate additional policy  preferences.

\begin{assumption}
\label{assump:zero_mean}
We assume that $r^\star \in\mathcal{R}$, where  
\begin{equation}
    \mathcal{R} = \bigg\{r:
    \exlim{\substack{x\sim\rho \\ y \sim \dnorm(\cdot| x)}}{r(x, y)} = 0.
    \bigg\}.
    \label{eq:zero_pi_mean}
\end{equation}
Here, $\rho$ is the prompt distribution and $\dnorm$ is a fixed calibration distribution independent of the algorithm.
\end{assumption}

\citet{cen2025vpo}  proposes a regularized MLE of the Bradley-Terry model \eqref{eq:MLE}, which  
appends a  bias term to the negative likelihood  
\begin{equation}
    r_{\mathsf{VPO}} = \arg\min_{r \in \mathcal{R} } \; \{\ell(r, \mathcal{D}) - \mathsf{sign} \cdot \alpha \cdot J^\star(r)\},
        \label{eq:RBMLE}
\end{equation}
incentivizing the algorithm to favor (resp. avoid) reward models with higher value  $J^\star(r) $ in the online (resp. offline) setting. Here, $\alpha > 0$ is a constant controlling the strength of regularization, and $\mathsf{sign}$ is set to $1$ in the online setting and $-1$ in the offline setting. 

At first glance, the objective function for value-incentivized policy optimization (VPO) \eqref{eq:RBMLE} does not immediately imply a computationally-efficient algorithm due to the presence of $J^{\star}(r)$. However, by exploiting the reward representation inferred from the closed-form optimal policy \eqref{eq:RLHF-policy-closed-form} via \eqref{eq:reward-policy-eq}, and Assumption~\ref{assump:zero_mean},
we can explicitly express $J^\star(r)$ as 
\begin{align}
    J^\star(r) 
    &= -\beta \exlim{ \substack{x \sim \rho \\ y \sim \dnorm(\cdot|x) }}{ \log \pi_r(y|x) - \log\pi_{\text{ref}}(y|x)}, \label{eq:Jstar}
\end{align}
where the second step follows because the bracketed term is independent of $y$ (c.f. \eqref{eq:RLHF-policy-closed-form}) and the last step follows from \eqref{eq:zero_pi_mean} whenever $r\in\mathcal{R}$.
Given this key ingredient, we can then rewrite \eqref{eq:RBMLE} to directly optimize the LLM policy, in a flavor similar to DPO, as
\begin{align}
  \pi_{\mathsf{VPO}} 
 &= \argmin_{\pi_r:\, r\in \mathcal{R} } \, \{\ell(r, \mathcal{D}) -  \mathsf{sign} \cdot \alpha \cdot J^\star(r)\}\notag\\
   &= \arg\min_{\pi} \Big\{-\sum_{(x^i, y_+^i, y_-^i)\in\mathcal{D}} {\log \sigma\Big(\beta\log\frac{\pi(y_+^{i} |x^{i})}{\pi_{\text{ref}}(y_+^{i}|x^{i})} - \beta\log\frac{\pi(y_-^{i}|x^{i})}{\pi_{\text{ref}}(y_-^{i}|x^{i})}\Big)} \notag\\
    &\qquad\qquad\qquad + \mathsf{sign} \cdot\alpha \beta  \exlim{ \substack{ x \sim \rho \\ y \sim \dnorm(\cdot|x)}}{\log \pi (y|x)  }\Big\},
    \label{eq:policy_RBMLE}
\end{align}
where we drop the constraint on $r\in \mathcal{R}$, since for any policy $\pi$ there exists $r\in\mathcal{R}$ such that $\pi =\pi_r$, and the last term without impacting the optimization solution. 
In fact, we can shift  the last term of \eqref{eq:policy_RBMLE} without impacting the optimization solution, recognizing that the regularizer amounts to adding an KL regularization 
$$\exlim{ \substack{x \sim \rho \\ y \sim \dnorm(\cdot|x)}}{\log \pi (y|x) - \log\dnorm(y|x)}\Big\} = - \exlim{x \sim \rho}{ \KL{\dnorm(\cdot|x)}{\pi(\cdot|x)}} $$  to the original DPO, which offers an interesting interpretation as pushing $\pi$ against/towards $\dnorm$ in the online/offline settings respectively, unveiling the role of reward calibration as a design choice in RLHF. 
 
We are now ready to instantiate the recipe of VPO to online and offline RLHF, respectively.


\paragraph{Online RLHF.} The online RLHF procedure extends training by performing reward learning and policy learning iteratively, with a growing preference dataset collected by using the current policy. We use $\pi^{(t)}$ to denote the policy used in the $t$-th iteration, where the superscript $^{(t)}$ indicates iteration $t$ in the online setting.
 The $t$-th iteration of VPO for online RLHF  consists of the following steps:
\begin{enumerate}
    \item \textit{New preference data generation.} We sample a new prompt $x^{(t)}\sim\rho$ and two answers $y_1^{(t)}, y_2^{(t)}\sim\pi^{(t)}(\cdot|x^{(t)})$, query the preference oracle and append $(x^{(t)}, y_+^{(t)}, y_-^{(t)})$ to the preference dataset.
    \item \textit{Policy update.} We train an updated policy with preference data $\mathcal{D}^{(t)} \coloneqq \{(x^{(s)}, y_+^{(s)}, y_-^{(s)})\}_{s=1}^{t}$ by minimizing the regularized negative log-likelihood, i.e., 
       \begin{align}   \label{eq:online_RBMLE}
   \pi^{(t+1)}                 &= \argmin_{\pi  } \Big\{-\sum_{s=1}^{t}{\log \sigma\Big(\beta\log\frac{\pi(y_+^{(s)}|x^{(s)})}{\pi_{\text{ref}}(y_+^{(s)}|x^{(s)})} - \beta\log\frac{\pi(y_-^{(s)}|x^{(s)})}{\pi_{\text{ref}}(y_-^{(s)}|x^{(s)})}\Big)} \nonumber \\
                &\qquad \qquad\qquad + \alpha \beta  \exlim{ \substack{ x \sim \rho \\ y \sim \dnorm(\cdot|x)}}{\log \pi(y|x)  }\Big\}.
            \end{align}
            
\end{enumerate}
Encouragingly, VPO admits appealing theoretical guarantees under linear function approximation. As established in \citet{cen2025vpo},   VPO --- without explicit uncertainty quantification --- achieves the same $\widetilde{O}(\sqrt{T})$ regret for online RLHF as its counterparts in standard contextual bandits with scalar rewards and using UCB for exploration \citep{lattimore2020bandit}.

%

\paragraph{Offline RLHF.}
In offline RLHF,  a fixed offline preference dataset is collected $\mathcal{D} \coloneqq \{x^{i}, y_+^{i}, y_-^{i}\}_{i=1}^{N}$, where $x^i\sim\rho$, $y^i\sim \pi_{\mathsf{b}}(\cdot|x)$ are sampled from a behavior policy $\pi_{\mathsf{b}}$, such as $\pi_{\text{sft}}$ from SFT. The proposed VPO for offline RLHF consists of one pass through the reward and policy learning phases, i.e.,
\begin{equation} \label{eq:RBMLE_offline}
    \widehat{r} = 
    \arg\min_{r\in\mathcal{R}} \; \{\ell(r, \mathcal{D}) + \alpha \cdot J^\star(r)\}
\quad   \mbox{and} \quad
    \widehat{\pi} = \arg\max_{\pi} J(\widehat{r}, \pi),
\end{equation}
which discourages over-optimization of the reward function given the limited offline preference data. 
In the same vein as deriving \eqref{eq:online_RBMLE}, and by leveraging \eqref{eq:Jstar}, we obtain the direct policy update rule:
\begin{align*} 
\widehat{\pi} & = \arg\min_{\pi} \Big\{-\sum_{i=1}^{N}{\log \sigma\Big(\beta\log\frac{\pi(y_+^{i}|x^{i})}{\pi_{\text{ref}}(y_+^{i}|x^{i})} - \beta\log\frac{\pi(y_-^{i}|x^{i})}{\pi_{\text{ref}}(y_-^{i}|x^{i})}\Big)} - \alpha \beta  \exlim{ \substack{x \sim \rho \\ y \sim \dnorm(\cdot|x)}}{\log \pi(y|x) }\Big\}.
\end{align*}
\citet{cen2025vpo} establishes the sub-optimality gap of VPO with linear function approximation  achieves the same rate of $\widetilde{\mathcal{O}}(1/\sqrt{N})$ as standard offline RL, as long as the offline dataset $\mathcal{D}$ has sufficient coverage.

\section{Concluding remarks}

In this tutorial, we have presented a small sample of the state-of-the-art advances in various RL settings and RL approaches, offering a fruitful interplay between statistical methodologies and RL algorithm development. 
We hope that this tutorial can equip new researchers with the core toolkits of modern RL, while motivating further theoretical and algorithmic investigations.

Before concluding, there are several remarks that we would like to make. 
First, while the algorithms introduced in this tutorial enjoy strong (often optimal) theoretical guarantees, many of them are designed under idealized mathematical assumptions. In practical large-scale RL applications---such as robotics, recommendation systems, and game playing---these conditions might not be met. As a result, practitioners often turn to alternative approaches that, while lacking optimal theoretical support, offer superior empirical performance in more complex scenarios. Partial examples include Deep Q-Networks and Proximal Policy Optimization, which have become de facto choices due to their scalability and adaptability and often outperform theoretically grounded algorithms when combined with domain-specific engineering and neural networks.  As another example, methods like Conservative Q-Learning or Advantage-Weighted Regression incorporate practical heuristics to cope with distributional shift and data coverage limitations, oftentimes even more effectively than the theoretically optimal approaches. All in all, while this tutorial emphasizes theoretical rigor and algorithmic optimality, it is important to recognize the gap between theory and practice. In fact, many modern RL systems integrate ideas from both ends of the spectrum, leveraging theoretical insights to inform  algorithm design.




Finally, we conclude by pointing out a few additional research directions that, due to space constraints, are not covered in this work but have recently attracted much attention. 
To begin with, while this tutorial has addressed sampling mechanisms like generative models, offline RL and online RL, 
it has omitted important settings in which an agent has simultaneous access to multiple sampling modalities. For example, an agent may be permitted to collect samples through real-time online exploration while also having access to an offline dataset. 
Such a hybrid RL setting has garnered significant recent attention from both theoretical and practical perspectives (e.g., \cite{xie2022role,song2023hybrid,li2024reward,wagenmaker2023leveraging,tan2024hybrid,tan2025actor}), offering insights that extend far beyond the scope of online RL or offline RL in isolation. In addition, an agent may have access to federated sampling modalities that can operate in parallel to accelerate learning, by marrying with advances in distributed optimization (e.g., \cite{woo2023blessing,woo2024federated,zheng2024federated,salgia2024sample,yang2024federated}).
Another important area pertains to  uncertainty quantification for RL algorithms, such as constructing valid and efficient confidence regions for value functions. In general, uncertainty quantification can be addressed  by establishing distributional guarantees (e.g., certain variations of central limit theorems) for the RL algorithms in use, which can become challenging in the presence of online data collection (see, e.g., \cite{samsonov2024improved,wu2024statistical,wu2025uncertainty}). Last but not least, we point out that the design and analysis of tractable RL algorithms with general function approximation remains an active area of research (e.g., \cite{foster2021statistical,wang2020reinforcement,agarwal2020flambe,chen2025unified,du2021bilinear,jin2021bellman,yang2025incentivize}), due to its instrumental role in modern practice to tame the enormous problem dimensionality.

\section*{Acknowledgement}
This work of Y.~Chi is supported in part by NSF CCF-2106778, DMS-2134080, AFRL FA8750-20-2-050 and ONR N00014-19-1-2404. 
Y.~Chen has been supported in part by the Alfred P. Sloan Research Fellowship, the Google Research Scholar Award, the AFOSR grants FA9550-19-1-0030 and FA9550-22-1-0198, the ONR grant N00014-22-1-2354, and the NSF grants CCF-2221009 and CCF-1907661.
Y.~Wei is supported in part by the Google Research Scholar Award, and the NSF grants CCF-2106778, DMS-2147546/2015447 and CAREER award DMS-2143215.
The authors are grateful to their collaborators for helpful discussions and contributions that shaped this tutorial, including but not limited to Changxiao Cai, Shicong Cen, Bo Dai, Simon Du, Jianqing Fan, Matthieu Geist, Jason Lee, Gen Li, Cong Ma, Yang Tong, Laixi Shi, Weichen Wu, Lin Xiao, Yuling Yan, Wenhao Zhan, and Zihan Zhang.

\def\newblock{\relax}%
\bibliographystyle{informs2014}
\bibliography{bibfileRL.bib,bibfileGame.bib,bibfileDRO.bib,RLHF_refs.bib,hybrid}

\end{document}